\documentclass[preprint,review,3p,times]{elsarticle}
\RequirePackage{tabularx}
\RequirePackage{multirow}
\RequirePackage{makecell}
\RequirePackage{subcaption}
\RequirePackage{graphicx}
\usepackage{amssymb}
\usepackage{amsmath}
\usepackage{booktabs}
\usepackage[numbers]{natbib}
\usepackage{xcolor}
\usepackage{xpatch}
 
\journal{Pattern Recognition}

\begin{document}
\begin{frontmatter}
% \newtheorem{definition}{Definition}
% \newcommand{\algorithmautorefname}{Algorithm}
% \newcommand{\eqautoref}[1]{\hyperref[#1]{Eq.~(\ref*{#1})}}
% \captionsetup[subfigure]{labelformat=simple}
% \renewcommand\thesubfigure{(\alph{subfigure})}
% \newtheorem{theorem}{Theorem} 
% \newtheorem{lemma}{Lemma}
% \newcommand{\lemmaautorefname}{Lemma}
% % \DeclareCaptionLabelFormat{myformat}{\textbf{Fig. #2}}
% % \captionsetup[figure]{labelformat=myformat}
% \def\figureautorefname{Fig.}


\title{Local and Global Feature Attention Fusion Network for Face Recognition}
\author{Wang Yu}
\ead{wyu981011@gmail.com}            
\author{Wei Wei}
\ead{weiw@hust.edu.cn}

\fntext[label2]{Wang Yu and Wei Wei was with Cognitive Computing and Intelligent Information Processing (CCIIP) Laboratory, School of Computer Science and Technology,Huazhong University of Science and Technology, Wuhan, China.}

\fntext[label4]{Wei Wei is the corresponding author.}

%% Abstract
\begin{abstract}
Recognition of low-quality face images remains a challenge due to invisible or deformation in partial facial regions. For low-quality images dominated by missing partial facial regions, local region similarity contributes more to face recognition (FR). Conversely, in cases dominated by local face deformation, excessive attention to local regions may lead to misjudgments, while global features exhibit better robustness. However, most of the existing FR methods neglect the bias in feature quality of low-quality images introduced by different factors. To address this issue, we propose a Local and Global Feature Attention Fusion (LGAF) network based on feature quality. The network adaptively allocates attention between local and global features according to feature quality and obtains more discriminative and high-quality face features through local and global information complementarity. In addition, to effectively obtain fine-grained information at various scales and increase the separability of facial features in high-dimensional space, we introduce a Multi-Head Multi-Scale Local Feature Extraction (MHMS) module. Experimental results demonstrate that the LGAF achieves the best average performance on $4$ validation sets (CFP-FP, CPLFW, AgeDB, and CALFW), and the performance on TinyFace and SCFace outperforms the state-of-the-art methods (SoTA).
\end{abstract}

% % \includegraphics{grabs}


%% Keywords
\begin{keyword}
Low-quality face recognition \sep Feature fusion


\end{keyword}

\end{frontmatter}




\section{Introduction}
\label{sec:intro}
Face recognition (FR) has gained widespread applications in many fields. However, effectively extracting identity information from low-quality face images remains challenging for face representation networks. 
Generally, low-quality face images are defined as those with extreme pose, face occlusion, poor contrast, and low resolution \cite{parde2016deep}. Key factors that influence FR performance include pose, lighting, occlusion, age, expression, etc \cite{trigueros2018face}.
Under such circumstances, the variability of face images may significantly limit the performance of FR. Therefore, the FR research of different head pose \cite{YANG2023HeadPose}, face occlusion \cite{HUANG2024110574}, various age \cite{Wang2024Age}, or low-resolution \cite{ZHANG2024Coupled} scenes has been widely concerned.

\begin{figure}[t]
  \centering
  \includegraphics[scale=0.7]{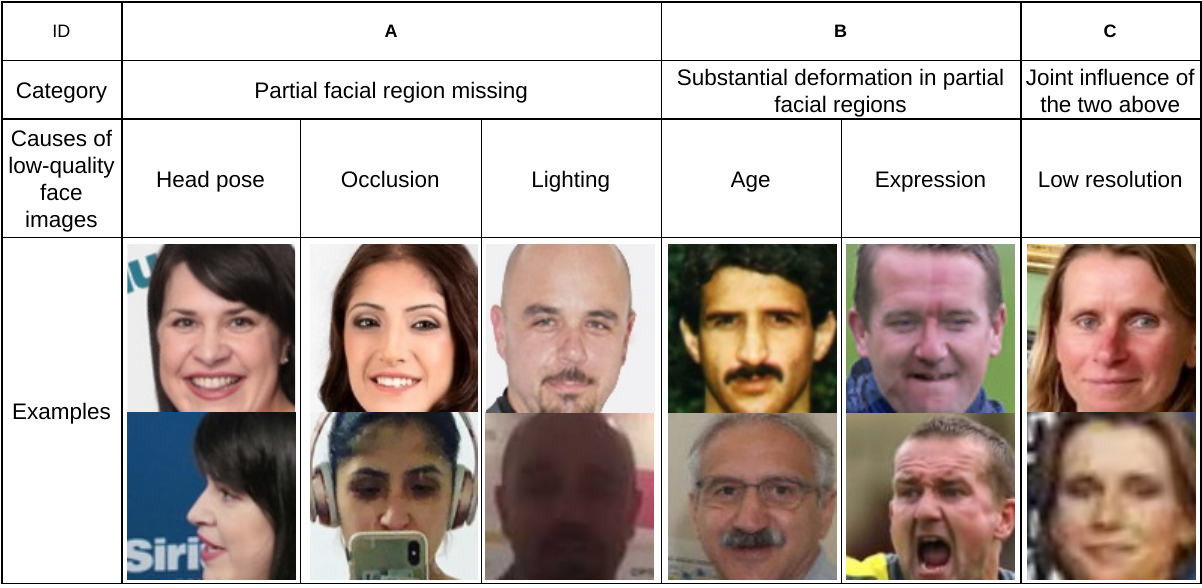}
\caption{Examples of low-quality face images caused by different factors. Each factor is illustrated using two contrasting face images. These factors generally increase the difficulty of FR.}
\label{fig:Factors influencing FR}
\end{figure}

To generalize further, we categorize the causes of low-quality face images into three main categories in reality (as depicted in Fig.\ref{fig:Factors influencing FR}): (A) Partial facial region missing. (B) Substantial deformation in partial facial regions. (C) Joint influence of the two above.
The factors resulting in partial facial region missing (Category A in Fig.\ref{fig:Factors influencing FR}) mainly stem from extreme head pose, facial occlusion, and brightness. This phenomenon leads to the attenuation of face structural information and increases the difficulty of FR.
Significant deformation occurs in partial facial regions (Category B in Fig.\ref{fig:Factors influencing FR}), such as expression and wrinkles caused by age changes, which will introduce noise to the local features. Excessive attention to these local regions often affects the performance of FR.
When missing and deformation jointly occur (Category C in Fig.\ref{fig:Factors influencing FR}), such as in low-resolution or with motion blur images, the attenuation of identity information is accompanied by noise, posing significant challenges for FR.

Recently, the main research directions of FR are divided into optimizing data preprocessing methods \cite{zhao2017dual,shi2021boosting}, optimizing loss functions \cite{CoReFace, HUANG2023PLFace} and optimizing feature representation networks \cite{wang2022cqa, Hybrid}. In this paper, we focus on the advancements of feature representation optimization in FR, which is mainly divided into two categories: global feature representation network and local feature representation network. With the development of deep neural networks, face representation networks based on global features have been widely applied due to their robustness. They primarily use classical CNN architectures directly or slightly modified as the backbone, such as ResNet\cite{he2016deep} and Mobilenet\cite{howard2019mobilenets}. The VPL model proposed by \citet{deng2021variational} focused on the sample-to-sample comparisons within a classification framework, enabling the model to further explore the optimal solution. Based on the Embedding Unmasking Model (EUM), \cite{Fadi2022Self} generates embeddings for faces in the presence of masks that resemble unmasked faces of the same identity.
After the vision transformers (ViTs) model was proposed, Transformer-based models have been gradually applied to FR. \citet{Hybrid} proposes a new plug-and-play hybrid token Transformer (HOTformer) module based on ViTs  to recognize key facial semantics by cooperation of atomic and holistic tokens. However, the face representation networks based on global features are vulnerable to partial facial region missing and background noise, which attenuates global facial information. As a result, they tend to suffer performance degradation in category A of Fig.\ref{fig:Factors influencing FR}.

Therefore, based on the idea that local patches play an important role in FR when the global face appearance changes significantly, some researchers have paid more attention to attention-based local feature representation networks in recent years. \citet{wang2020hierarchical} introduced the HPDA networks, designed to extract facial features of different local regions at various scales. The CQA-Face proposed by \citet{wang2022cqa} 
emphasizes non-critical but still discriminative local regions to comprehensively explore useful facial parts and suppress noisy parts in a global scope.
Although attention-based local feature representation networks assist models in capturing more effective face information under extreme pose or occlusion (category A in Fig.\ref{fig:Factors influencing FR}), their reliance on partial facial region information makes them more vulnerable when local regions undergo deformation, such as expression or age change (category B in Fig.\ref{fig:Factors influencing FR}) than global feature extraction networks.

Observing the FR process of low-quality facial images under different categories, we find that when samples are primarily affected by missing facial regions (category A in Fig.\ref{fig:Factors influencing FR}), the global structural information decays rapidly with the deepening of the missing degree, and local information is more stable than global information due to focusing only on specific facial areas. Consequently, people tend to prioritize identity recognition through local region similarity. When the samples have significant facial region deformation (category B in Fig.\ref{fig:Factors influencing FR}), the local face information is attenuated due to noise, while global information is more conducive to recognition. We define the effectiveness of features for FR tasks as feature quality. Low-quality images caused by different categories have biases in local and global feature quality. 

However, the existing FR models have not considered the feature quality bias nor designed an effective method to utilize the characteristics to help the model obtain more effective facial information. 
To solve this issue, we introduce the Local and Global Feature Fusion (LGF) module. This module drives feature norm as a proxy for feature quality to dynamically measure the attention between local and global features at a relatively low cost.
When the structural information of global features is weakened due to incomplete facial regions, the LGF module prioritizes effective local face information to enhance face features. Conversely, when local features introduce noise due to deformation in certain face areas, more attention is directed towards global features with greater robustness to complement facial information and improve overall feature quality.

In addition, we propose a Multi-Head Multi-Scale Local Feature Extraction (MHMS) module. Assuming the distribution of local information at different scales across various spatial positions, we calculate spatial attention for multiple scales respectively to obtain information distributed across various visual fields. Channel attention is implemented to enhance local information and suppress noise. In addition, the multi-head structure ensures that the model can obtain sufficiently rich local face features, enhancing the separability of identity information in high-dimensional space.
By connecting the MHMS, the Global Feature Extraction (GFE) module, and the LGF, we develop a new face recognition network architecture: the Local and Global Feature Attention Fusion (LGAF) network. Our main contributions are threefold:

\begin{itemize}
    \item A Local and Global Feature Fusion (LGF) module is proposed. We explore the correlation between the feature norm and the three categories of causes for low-quality face images (as shown in Fig.\ref{fig:Factors influencing FR}) and drive the feature norm as a proxy for feature quality. Exploiting the information of local and global features quality, the module dynamically measures the attention between local and global features. This realizes the information complementarity between local and global features and reduces feature quality biases introduced by different categories of low-quality images. In particular, the LGF does not add excessive cost to the network. To our knowledge, this is the first attempt in FR to fuse local and global features based on attention. The LGF can help the model get more discriminative and high-quality face features.

    \item A Multi-Head Multi-Scale Local Feature Extraction (MHMS) module is proposed. The module can dynamically evaluate the significance of various facial regions across spatial and channel dimensions at multiple scales, thereby exploring more comprehensive local face information. When partial facial regions are missing, the multi-head structure of the MHMS ensures that the model extracts enough face information to ensure recognition stability.

    \item We verify the efficacy of the LGAF on $6$ datasets (CFP-FP, CPLFW, AgeDB, CALFW, TinyFace, and SCFace). The results demonstrate that the LGAF network outperforms SoTA on two low-resolution face datasets, and achieves the best average performance on $4$ high-resolution face datasets.
\end{itemize}

\section{Related Work}
\label{sec:Related Work}
%-------------------------------------------------------------------------
\subsection{Local Feature Extraction}

Attention-based local feature representation networks in FR have gradually attracted the attention of the academic community. The Local and multi-Scale Convolutional Neural Networks (LS-CNN) was proposed by \citet{wang2019ls}, which is one of the early works that introduced attention to face representation. The Hierarchical Pyramid Diverse Attention (HPDA) network \cite{wang2020hierarchical} used a Pyramid Diverse Attention (PDA) to extract face features of different scales and utilized a Hierarchical Bilinear Pooling (HBP) to fuse information from multiple scales. However, the PDA with different-size pooling operations may lose some fine-grained features. The HBP did not consider the effectiveness difference of various scale features and introduced redundant information. The CQA-Face \cite{wang2022cqa} extracted more local face features that are not the most important but still discriminative. When the key region is occluded, it can also rely on the features of other regions for FR. However, it did not account for identity information that may appear dynamically at different scales. To solve the problems, we propose the MHMS module. It uses convolution kernels at various scales to simultaneously obtain local feature maps in different visual fields and implements spatial attention emphasis on effective regions for multiscale feature maps, respectively. The Squeeze-and-Excitation (SE) module \cite{hu2018squeeze} in the MHMS is utilized to measure which scales of local features the model should focus on. We employ a multi-head structure to help the model obtain more local information and enhance the separability in high-dimensional space.

In addition, the existing local feature representation networks are often suitable for FR of low-quality images caused by partial facial region missing (facial occlusion, extreme pose, etc.). When the samples have large local face deformation, these methods may cause mismatching. Therefore, we propose the LGF module to reduce the feature quality biases caused by different low-quality image causes, as detailed in Sec.\ref{LGF}

%-------------------------------------------------------------------------
\subsection{Feature Norm}

\citet{parde2016deep} proved that face features based on convolutional neural networks retain face pose information. \citet{ranjan2017l2} proposed for the first time that there was a certain relationship between the L2-norm of face features and face image quality. \citet{meng2021magface} used feature norm as an indicator to measure the quality of the given face image, and implemented fine-grained constraints within the class. \citet{kim2022adaface} found that the feature norm has the characteristics of fast convergence, long-term stability, and a high correlation with image quality (IQ) score.
In contrast to our elaborate categorization of the factors influencing the quality of facial features (such as posture, age variation, and so on), image quality in the AdaFace is a combination of attributes that indicates how faithfully an image captures the original scene.
Based on the idea that face image quality affects the difficulty of sample recognition, the AdaFace used the feature norm as a proxy to make the model pay more attention to the samples with low image quality. 

However, the AdaFace only explores the correlation between feature norm and IQ score. To further explore the properties of feature norm, we calculate and analyze the trends in feature norm variations caused by partial facial region missing, deformation in partial facial regions, and joint influence of the two above, resulting in low-quality facial images, as detailed in Sec.\ref{feature norm}.

\section{Methods}

\label{sec:Methods}
\begin{figure*}[t]
\centering
\includegraphics[scale=0.7]{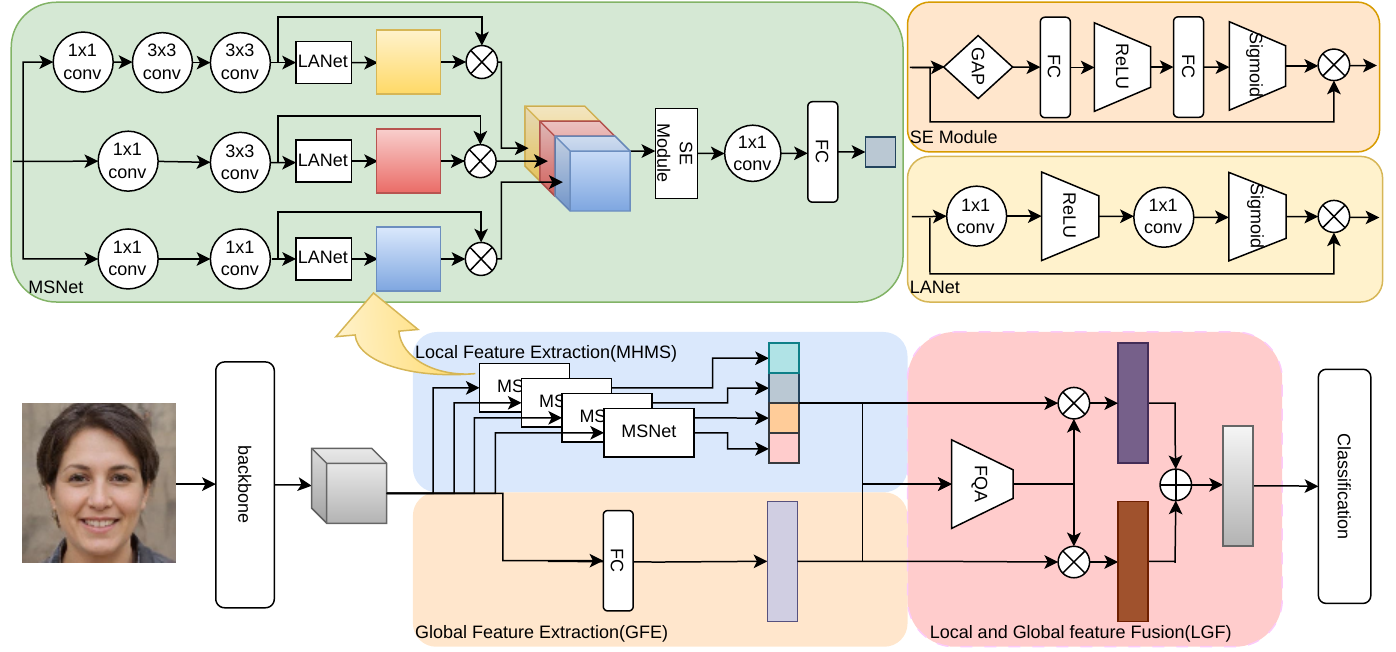}
\caption{The framework of the proposed Local and Global Feature Attention Fusion (LGAF) network. GAP represents a global average pooling operation, $a\times a$ conv represents a convolution operation with kernel size $a$, FC stands for a fully connected layer, and FQA represents a feature quality assessment module. The MHMS and GFE are responsible for extracting effective local features and global features, respectively. The LGF module adaptively allocates attention between local and global features and performs feature fusion. The MSNet utilizes LANet and SE Module to extract local features at various scales.}
\label{fig:The overall network structure of LGAF}
\end{figure*}

The LGAF model comprises five key components: the backbone network, the local feature extraction module, the global feature extraction module, the feature fusion module, and the classification layer. The ResNet100 is selected as the backbone to extract face feature maps $C(x_i)$ from processed face images $x_{i} \in X^{w \times h \times c}$ where $h$, $w$, and $c$ refer to the height, width, and number of channels, respectively. 
The local feature extraction module adopts a Multi-Head Multi-Scale Local Feature Extraction (MHMS) module, composed of several MSNet networks. A single MSNet implements spatial attention at each scale to obtain more efficient local information. Additionally, it utilizes channel attention to minimize redundancy and enhance the representation. The multi-head structure formed by multiple MSNet modules ensures that the model comprehensively captures rich facial information.
The Global Feature Extraction (GFE) module transforms the face feature map $C(x_i)$ into a compact and discriminative global feature $\Upsilon_i$ through linear transformation, where $\Upsilon_i=G(C(x_i))$ and $G(\cdot)$ represents a projection, implemented with a fully connected layer. To dynamically measure the attention between local and global features, we propose a Local and Global Feature Fusion (LGF) module. The LGF module aims to achieve information complementation between local and global features to reduce feature quality biases. The network structure of the LGAF is shown in Fig.\ref{fig:The overall network structure of LGAF}.

\subsection{Feature Norm}
\label{feature norm}
Although previous work has revealed partial properties of feature norm, it is not comprehensive. To further explore the relationship between feature norm and the causes of low-quality images, we designed three experiments. Considering the reliability and convenient access of the data source, we use the data of pose variation, expression variation, and different motion blur intensity to represent the low-quality face data mainly caused by partial facial region missing, deformation in partial facial regions, and joint influence of the two above (as shown in Fig.\ref{fig:Factors influencing FR}).

\subsubsection{Pose and feature norm}
Employing head pose as a controlled variable, we focus on the relationship between head yaw angles and feature norm, which represents the relationship between feature norm and the degree of facial region missing. We select face images under natural light from the Multi-PIE dataset \cite{gross2010multi} to explore the variation trend between feature norm and the partial facial regions missing caused by pose variation.

\begin{figure}[!t]
  \begin{subfigure}{1\linewidth}
    \centering
\includegraphics[scale=0.6]{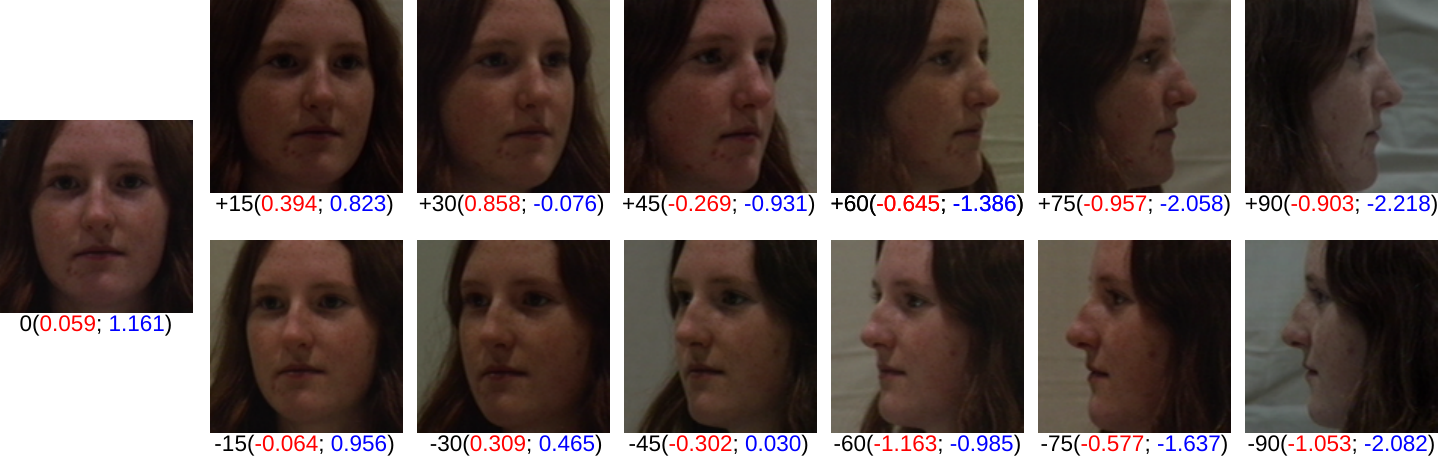}
\subcaption{}
\label{fig:Pose}
  \end{subfigure}
  \hfill
    \begin{subfigure}{1\linewidth}
    \centering
\includegraphics[scale=0.8]{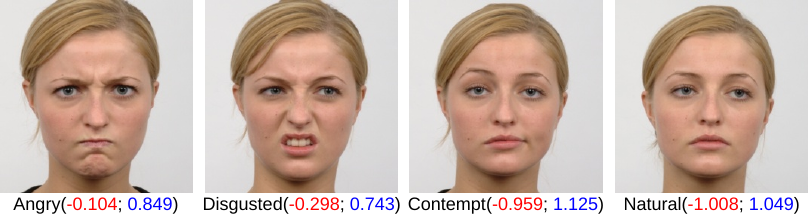}
\subcaption{}
\label{fig:expression}
  \end{subfigure}
      \begin{subfigure}{1\linewidth}
      \centering
  \includegraphics[scale=0.8]{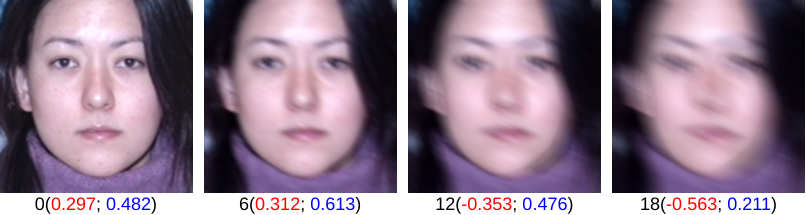}
\subcaption{}
\label{fig:motion blur}
  \end{subfigure}
\caption{(a) Examples of feature norms at different yaw angles. Comments below images represent yaw angle (\textcolor{red}{local feature norm}; \textcolor{blue}{global feature norm}).  (b) Examples of feature norms at different expression. Comments below images represent expression (\textcolor{red}{local feature norm}; \textcolor{blue}{global feature norm}). (c) Examples of feature norm under different motion blur intensities. Comments below images represent moving interval pixel length (\textcolor{red}{local feature norm}; \textcolor{blue}{global feature norm}).}

  \label{fig:Pose and Norm}
\end{figure}

\begin{figure}[!th]
  \centering
    \begin{subfigure}{0.49\linewidth}
\includegraphics[scale=0.5]{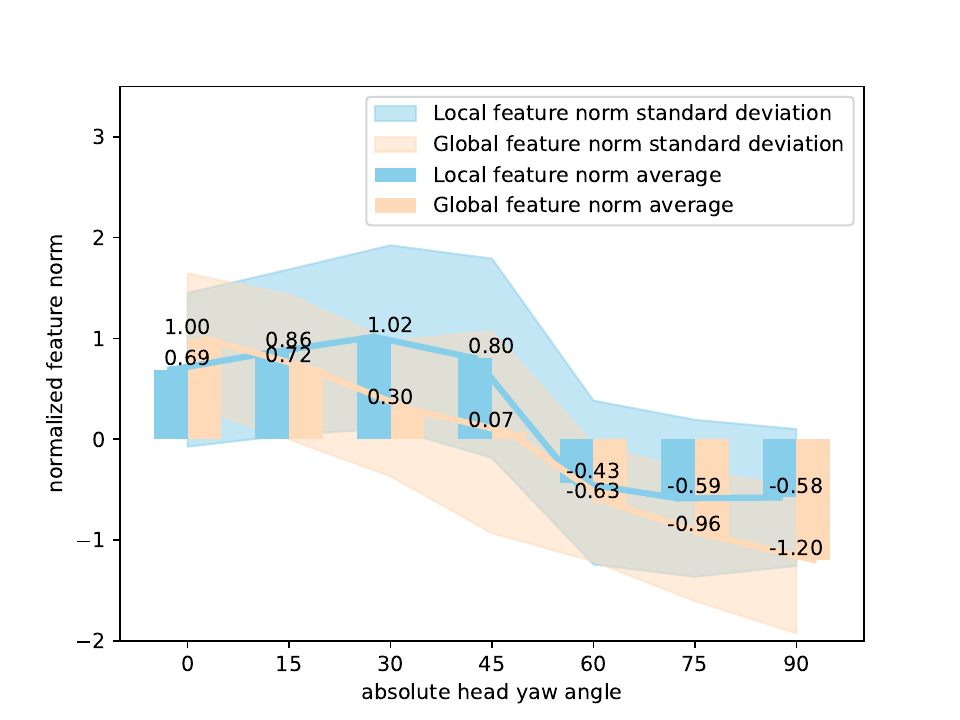}
\caption{}
\label{fig:Pose analyse}
  \end{subfigure}
      \begin{subfigure}{0.49\linewidth}
  \includegraphics[scale=0.5]{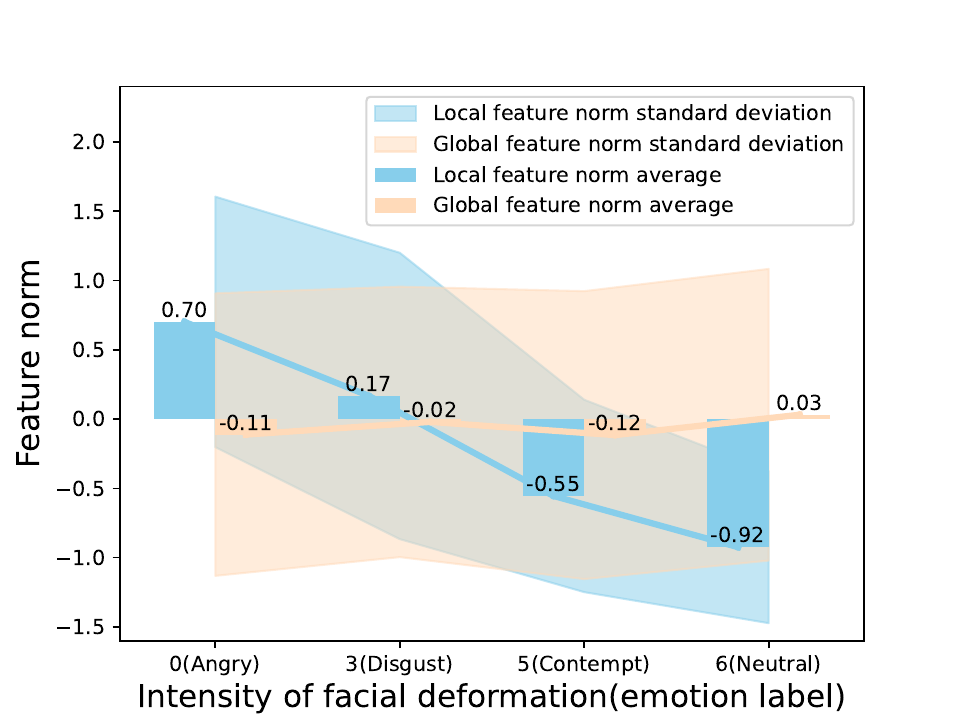}
\caption{}
\label{fig:expression analyse}
  \end{subfigure}
      \begin{subfigure}{0.49\linewidth}
  \includegraphics[scale=0.5]{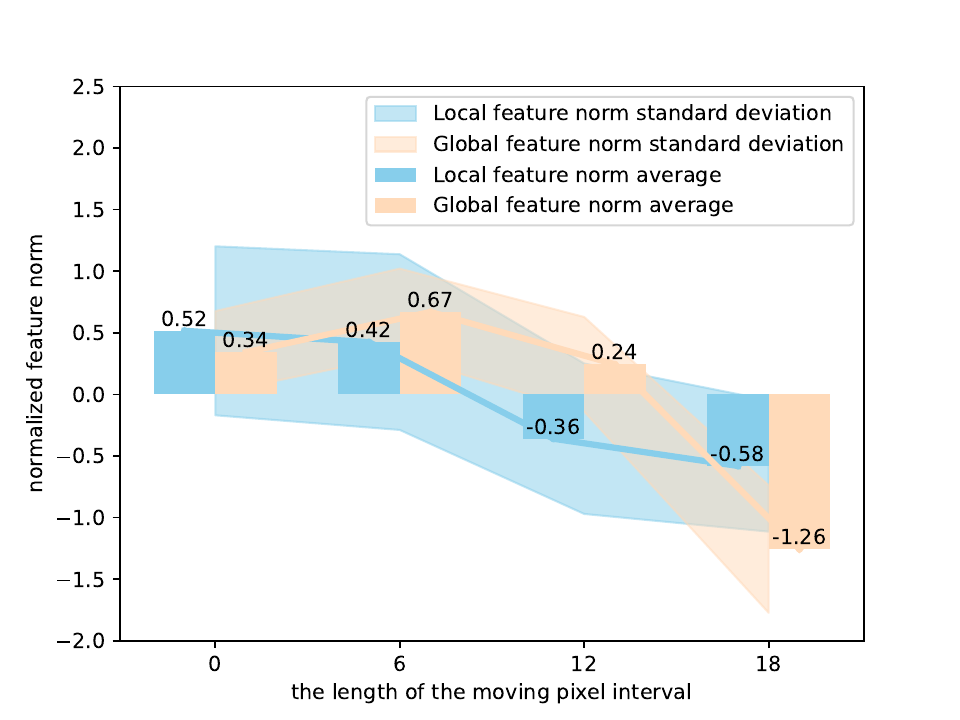}
\caption{}
\label{fig:motion blur analyse}
  \end{subfigure}
      \begin{subfigure}{0.49\linewidth}
  \includegraphics[scale=0.5]{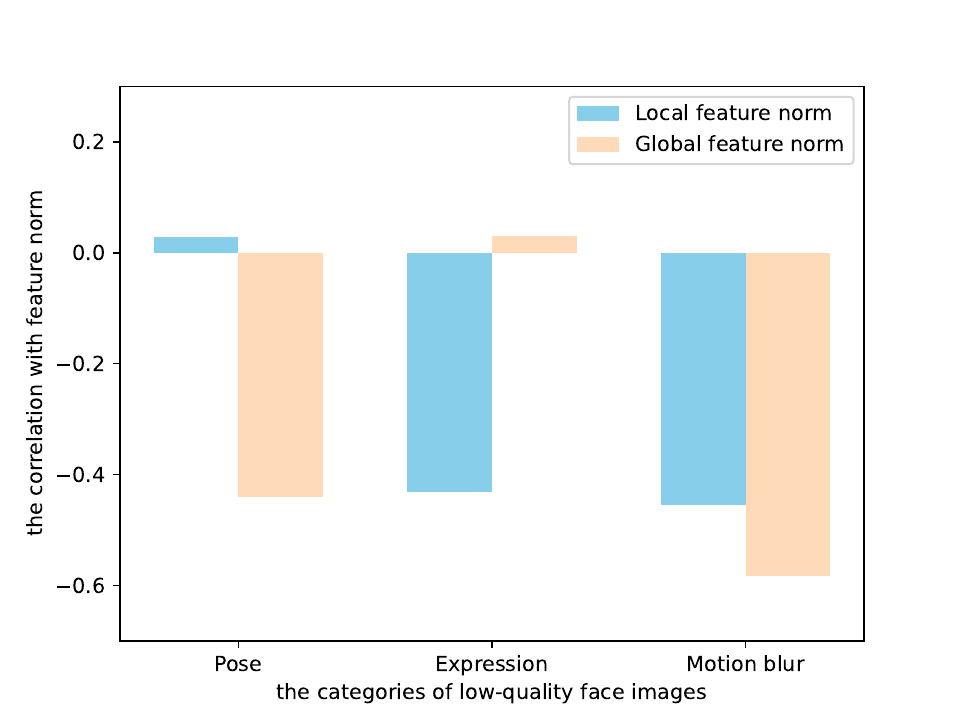}
\caption{}
\label{fig:correlation}
  \end{subfigure}

  \caption{Examples of the mean and standard deviation of the feature norm under different poses, ages, and motion blur intensities, as well as the correlation scores between feature norm and pose, age, and motion blur. (a) Local and global average feature norm at absolute yaw angle between 0 and 90 degrees. (b) Local and global average feature norm under different expression. (c) Local and global average feature norm under varying motion blur intensities.
 (d) The correlation scores between influencing factors of low-quality face images and feature norm.}
  \label{fig:age, blur and correlation}
\end{figure}

As shown in Fig.\ref{fig:Pose}, the larger the head yaw angle, the larger the facial missing area. To mitigate individual differences affecting the experimental results, we calculate the global and local average feature norm of different absolute yaw angles (Fig.\ref{fig:Pose analyse}). When the absolute yaw angle is in the range of 0 to 45 degrees and 60 to 90 degrees, the local feature norm is relatively stable and does not change significantly with the expansion of the face region missing. When the absolute yaw angle is between 45 and 60 degrees, almost half of the face information is lost, and the local feature norm decreases greatly. In contrast, the global feature norm nearly decreases linearly with the change of absolute yaw angle. The experiment shows that the local feature norm is relatively stable for partial face region missing within a certain range, while the global feature norm is easily affected.
As shown in Fig.\ref{fig:correlation}, we analyze the correlation between the absolute head yaw angle within 0 to 45 degrees and the norm of local and global features, yielding correlation scores of 0.028 and -0.441, respectively.

\subsubsection{Expression and feature norm}
We illustrated the relationship between facial local deformation intensity and feature norm by using the relationship between different expression and feature norm. From the RaFD dataset \cite{langner2010presentation}, we selected facial data with natural, disgust and contempt labels. Contempt, disgust and angry are expressed as:
\begin{equation}
Contempt=AU14,
\label{eq:Contempt}
\end{equation}
\begin{equation}
Disgust=AU9+AU10+AU25,
\label{eq:Disgust}
\end{equation}
\begin{equation}
Angry=AU4+AU5+AU7+AU17+AU23+AU24,
\label{eq:Angry}
\end{equation}
where AU4, AU5, AU7, AU9, AU10, AU14, AU17, AU23, AU24 and AU25 correspond respectively to brow lower, upper lid raise, lid tighten, nose wrinkle, upper lip raise, dimple, chin raise, lip tighten, lip press and lips part. Based on the number of action units involved, we consider three different expression as three intensities of facial distortion changes, in order: angry, disgust, contempt and natural (Fig.\ref{fig:expression}). As shown in Fig.\ref{fig:expression analyse}, the correlation score between facial local distortion intensity and local feature norm is -0.43, while the correlation score between facial local distortion intensity and global feature norms is 0.03 The experiments indicate that as the intensity of facial distortion increases, there is an almost linear decline in local feature norm, while global feature norm remains relatively stable.

\subsubsection{Motion blur and feature norm}
It is observed that motion blur is one of the main causes of low-quality face images. The increase of motion blur intensity may not only lead to substantial deformation in partial facial regions (such as the eye shape changes at Length 12 in Fig.\ref{fig:motion blur}), but also result in the loss of partial facial regions (such as the eyebrow region information missing at Length 18 in Fig.\ref{fig:motion blur}). We generate motion-blurred images from face images of the frontal face under flash light in the Multi-PIE dataset \cite{gross2010multi}. As shown in Fig.\ref{fig:motion blur analyse}, we use moving interval pixel lengths of 0, 6, 12, and 18 representing multiple blurring levels. As motion blur intensity increases, local and global feature norm nearly exhibit a decreasing trend, and their correlation scores are -0.456 and -0.584, respectively (as depicted in Fig.\ref{fig:correlation}).

The above three experiments prove that the feature norm can serve as a proxy to evaluate local and global feature quality. (a) In cases of low-quality samples primarily caused by missing local facial regions (e.g., extreme pose, occlusion, etc., as depicted in category A of Fig.\ref{fig:Factors influencing FR}), local features are more conducive to recognition due to the focus on local similarity, while global features deteriorate with information loss. The global feature norm also decreases with the expansion of the missing degree, and the local feature norm remains relatively stable. (b) For samples affected only by deformation in partial facial regions (e.g., expression, age change, makeup, etc., as shown in category B in Fig.\ref{fig:Factors influencing FR}), excessive emphasis on local features may lead to mismatching, while global features, with better robustness, are more favorable for identity recognition. As the intensity of deformation increases within local face regions, there is a clear response from the local feature norm, while the global feature norm remains relatively stable. (c) When the samples are simultaneously impacted by missing and deformation within partial facial regions (such as low-resolution images shown in category C of Fig.\ref{fig:Factors influencing FR}), the quality of local and global features decreases due to high variability, and there is a high correlation between feature norm and motion blur intensity. Therefore, we treat the feature norm as a proxy for local and global feature quality to help the model measure feature quality within the LGF module.

\begin{figure}[!t]
\centering
\includegraphics[scale=0.45]{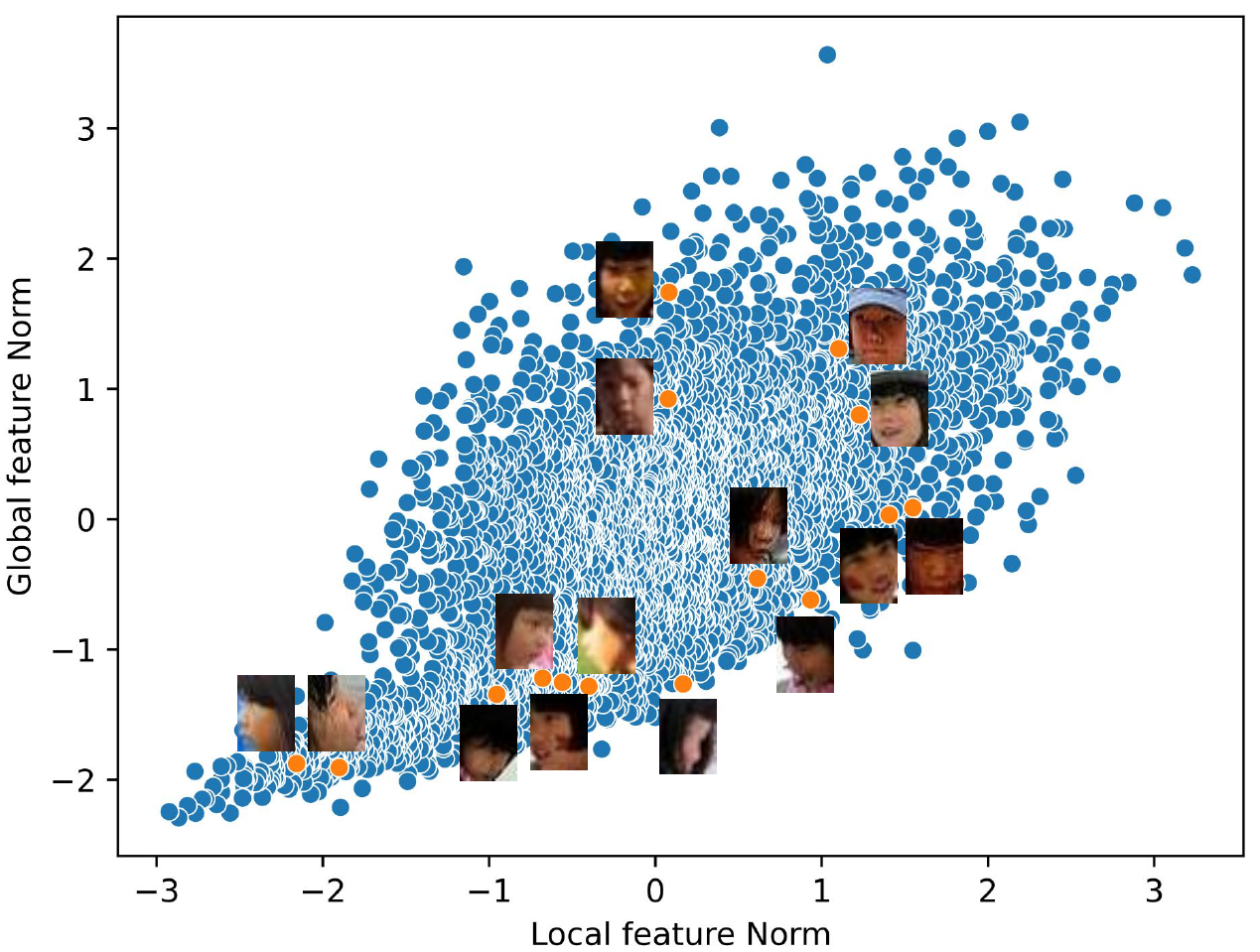}
\caption{Feature norm distribution statistics and partial sample visualization on TinyFace.}
\label{fig:norm tinyface}
\end{figure}

Furthermore, we conduct a statistical analysis of the feature norm distribution on TinyFace dataset, which is a low-quality dataset reflecting real-world unconstrained scenarios. In Fig.\ref{fig:norm tinyface}, we visually display a portion of the samples, revealing that many faces with low global feature norm are notably absent. Additionally, face images with low local feature norm exhibit varying degrees of facial deformation. These results demonstrate that the correlation between the causes of low-quality images in the real world and their local and global feature norm further validates the use of feature norm as an indicator of feature quality.

\subsection{Local and Global Feature Fusion}
\label{LGF}

For low-quality face images caused by different factors, local features and global features have their own advantages and disadvantages. Driven by face feature quality, we propose a Local and Global Feature Fusion (LGF) module to realize effective information complementation between local and global features without excessive cost to the network. However, the selection of facial feature quality evaluation metric plays a key role. We consider a simple and reliable way to measure the quality of feature embeddings. It is assumed that the influence of partial face regions missing and deformation on the sample are independent of each other. When the sample is missing part of the face region, as the missing region expands, the method should be sensitive to the global features while maintaining stability concerning local features. When the sample is mainly affected by the local deformation of the face, with the deepening of the local deformation, the method should display sensitivity towards local features, but stable to global features. 

We measure the local or global feature energy to evaluate the effective amount of information contained in a feature, and then use it as a proxy for feature quality. The formula for calculating the feature energy is as follows:
\begin{eqnarray}
    E(T^j) &=& \Vert T^j \Vert \\
                    &=& \sqrt{\sum_{i=1}^n (\tau_i^j)^2},
                \qquad j \in (l,g),
    \label{eq:energy}
\end{eqnarray}

Where $E(\cdot)$ denotes the feature energy, $T^l=\Upsilon$ represents the local feature, $T^g=\Psi$ represents the global feature, and $\tau_i$ denotes the components of $T$.
In the calculation form, the feature energy has the same calculation form as the feature norm, considering alignment with previous research, it is uniformly referred to as the feature norm in this paper.
Through the experiments in Sec.\ref{feature norm}, it is proved that the feature norm has a similar change trend and can be used as a proxy of the feature quality.

Local feature norm $Z_i^l$ and global feature norm $Z_i^g$ are calculated by the Feature Quality Assessment (FQA) module, where $Z_i^l=\Vert\Upsilon_i\Vert$ and $Z_i^g=\Vert\Psi_i\Vert$. We normalize the feature norm using the batch statistics  $\mu_Z^j$ and $\sigma_Z^j$ as follows:
\newcommand{\clip}[1]{\left\lfloor#1\right\rceil}
\begin{equation}
    \widehat{Z_i^j} = \clip{ h * \frac{Z_i^j - \mu_{Z}^j}{\sigma_{Z}^j}}^1_{-1}+1,\qquad j \in (l,g),
    \label{eq:norm}
\end{equation}
where $\clip{}^1_{-1}$  clips the value to [-1, 1], and $\widehat{Z_i^j} \in [0, 2]$. The distribution of $\frac{Z_i^j - \mu_{Z}^j}{\sigma_{Z}^j}$ is approximately unit Gaussian, and we use a hyperparameter $h$ to regulate the distribution concentration. Following the 3-sigma rule based on the unit Gaussian distribution, we determine that setting the hyperparameter $h$ to 0.333 ensures that the vast majority of samples fall within the range of -1 to 1. Batch statistics are updated by the exponential moving average (EMA) to ensure stability. $t$ indicates $t$-th step, then
\begin{equation}
    \mu_Z = \alpha \mu_Z^{(t)} + (1-\alpha) \mu_Z^{(t-1)},
\end{equation} 
\begin{equation}
\sigma_Z = \alpha \sigma_Z^{(t)} + (1-\alpha) \sigma_Z^{(t-1)}, 
\end{equation}
where $\alpha$ is set to 0.01 for momentum. 

The attention between local and global features is calculated as follows:
% \begin{equation}
% \gamma_i^l,\gamma_i^g=M(\Omega( \widehat{Z_i^l}, \widehat{Z_i^g})),
% \label{eq:attention}
% \end{equation}

\begin{equation}
\gamma_i^j=\frac{\widehat{Z_i^j}}{\sum_{j \in (l,g)}\widehat{Z_i^j}},
\label{eq:attention}
\end{equation}
and $\gamma_i^j$ represents the attention weight for features. When partial facial region missing is the dominant factor of low image quality, the model will allocate more attention to local features to keep the recognition performance stable. When substantial deformation in partial facial regions leads to low-quality images, the model will allocate more attention to global features to avoid excessive focus on local similarities that may result in mismatching.

The new face feature after the LGF module is shown as follows:
\begin{equation}
\kappa_i=\gamma_i^l\Upsilon_i+\gamma_i^g\Psi_i,
\label{eq:fusion feature}
\end{equation}
$\kappa_i$ as a new face feature is normalized and fed to the classification layer to calculate the probability of each class. Through the LGF module, the model dynamically allocates the attention weight between local and global features, which is beneficial to obtain more discriminative and high-quality features to help the model achieve stable recognition performance.

\subsection{Multi-Head Multi-Scale Local Feature Extraction Module}

Existing methods for local feature extraction lack effectiveness in obtaining and enhancing local information across various scales. Inspired by \cite{szegedy2015going}, we propose a Multi-Scale Local Feature Extraction Network (MSNet), which utilizes multi-scale convolution operations to capture local features. Compared to the LS-CNN network, we believe that local features at different scales have distinct spatial positional distributions. Therefore, while allowing the MSNet module to comprehensively consider information from various range neighboring locations during attention computation, we extract spatial attention for local feature maps at each scale to emphasize crucial local regions. Moreover, the SE module is employed to implement channel attention for filtering redundant information. By leveraging spatial and channel dual-dimensional attention, it achieves a thorough exploration of identity information at various scales. Considering that the information of key regions cannot always be effectively obtained (e.g. key parts are occluded), a single local feature is vulnerable. To address this issue, we adopt the multi-head structure for the MHMS module to extract multiple local features simultaneously. It helps the model maintain stable recognition performance and increases separability in high-dimensional space. Even when the facial features of some regions are difficult to extract, the local information of the remaining regions can still help the model for identity recognition.

We use convolution operations of various kernel sizes to extract the face feature maps given by the backbone network. Based on the idea that local features at different scales have varying spatial distributions, we measure the spatial attention at each scale using the LANet proposed by \citet{wang2019ls}. The LANet consists of two layers of 1x1 convolution operations for calculating local spatial attention. Regions judged as important by the LANet are assigned higher attention, while less discriminative regions receive lower attention. Through this module, we can focus on the local regions that are beneficial to recognition and reduce the noise brought by less relevant regions. Let $t$ be set to 1, $P(\cdot)$ be a LANet, and $GAP(\cdot)$ denote a global average pooling operation. $P(\cdot)$ is defined as follows:
\begin{equation}
P(\omega)=\omega \Phi_t(\Phi_t(GAP(\omega))),
\label{eq:LANet}
\end{equation}
and the local feature maps obtained after implementing spatial attention at each scale are represented as follows: 
\begin{equation}
l^j_{i}=P(\Phi_{j}(C(x_i))),
\label{eq:local feature maps}
\end{equation}
where $\Phi_j(\cdot)$ is the convolution operation to obtain local feature maps at various scales and $\Phi_j(\omega)=W_j\omega$. Specifically, j represents the size of the equivalent convolution kernel.

Since the key local information of FR is scattered at different scales, local face features are highly discriminative at some scales, but local information at other scales may lead to confusion. Therefore, concatenating them directly may lead to performance degradation. We adopt the SE module to apply channel attention to the feature maps of different scales, which will enhance the effective representation of local features. $S(\cdot)$ represents the SE module and the local feature provided by the $k$-th MSNet module is as follows:
\begin{equation}
L^k_{i}=F(S(\Omega(l^1_{i},...,l^j_{i}))),
\label{eq:MS local feature}
\end{equation}
where $\Omega(\cdot)$ represents a concatenation operation and $F(\cdot)$ represents a projection of local features. 

A single local feature is vulnerable to partial facial region missing. Therefore, we adopt a multi-head structure to extract $b$ local features simultaneously. When effective local features cannot be extracted from certain regions, the remaining local features can still contribute to stable recognition. The local features obtained by the MHMS are as follows:
\begin{equation}
\Psi_i=\Omega(L^1_i,...,L^b_i), 
\label{eq:local feature}
\end{equation}

\section{Experiments}
\label{sec:Experiments}

\subsection{Datasets and Implementation Details}
\subsubsection{Training set} 
We use MS1MV2 \cite{deng2019arcface} and WebFace4M \cite{zhu2023webface260m} as our training datasets. MS1MV2 has a total of 5.8M face images in 85K individuals by cleaning noise labels from MS-Celeb-1M \cite{guo2016ms} dataset. WebFace4M is a dataset comprising 10\% of the data from WebFace260M, and it contains a total of 206K individuals and 4.2 million face images.

\subsubsection{Test set}
We select CFP-FP \cite{sengupta2016frontal}, CPLFW \cite{cplfw}, AgeDB \cite{agedb}, and CALFW \cite{calfw}, which are widely utilized FR benchmarks for 1:1 verification, to assess the recognition performance of the models in low-quality scenes caused by diverse factors.
The CFP-FP dataset is designed to gather face images under various poses in unconstrained scenarios. Each identity encompasses 10 frontal face images and 4 side faces with distinct poses, aiming to evaluate the performance of face verification in the real world.
The CPLFW dataset collects 3,000 positive face pairs from the Internet with pose differences while minimizing other attribute disparities for studying the issue of pose invariance in FR.
The AgeDB dataset collects face images of celebrities of different ages via the Internet. The dataset incorporates a considerable number of face images of varying ages, ranging from 1 to 101 years old.
The CALFW dataset is a validation set of faces that vary with age in unconstrained scenarios and focuses on those affected by age variability in FR.

TinyFace \cite{cheng2019low} is a low-resolution face dataset collected from public web data for 1: N recognition tests. The images were collected under circumstances of uncontrolled pose, illumination, occlusion, and background. SCface \cite{2011SCface} was captured in an uncontrolled indoor environment, and images from commercial cameras with different resolutions simulate real-world conditions, concentrating on different surveillance use case scenarios.

\subsubsection{Training setting}
We resize the original pictures into 112x112 images and apply the same data augmentation strategy as in \cite{kim2022adaface}. Random rectangle cropping, photometric augmentation, and rescaling operations are applied with a probability of 0.2. These operations expand the quantity of low-quality images in the training dataset. Random rectangular cropping and photometric augmentation enhance the impact of the partial facial region missing from the original image. Image rescaling involves resizing the image to a smaller size and back, resulting in blurriness. This operation simulates the low-quality images caused by low resolution. ResNet100 is selected as the backbone network, and ArcFace loss and CosFace loss as the loss functions to evaluate the performance of the LGAF, respectively. For loss functions, the scale parameter $s$ is 64, and the margin parameter $m$ is set to 0.4. The model is trained using the SGD optimizer with an initial learning rate of 0.1 and step scheduling at 12, 20, and 24 epochs. 

\subsection{Ablation and Analysis}
\label{different fusion modes}

We use ResNet100 and MS1MV2 as the backbone and training set for ablation experiments, and the ArcFace loss as a default loss function. 
% The average performance of four 1:1 face verification sets is used to be a performance metric. 
The ablation experiments investigate the following three aspects: (a) Effects of the MHMS, the GFE, and the LGF; (b) The settings of parameter $b$; (c) Effects of different fusion modes in the LGF.

\begin{table}[th!]
\centering
\caption{Performance comparison between the MHMS module, the GFE module, and the LGF module. When only global features are extracted, the model degenerates to ArcFace. $\checkmark$= the module is used and *=our evaluation of the released model.}
\begin{subtable}[t]{1\linewidth}
    \centering
    \subcaption{The evaluation is performed on four 1:1 validation sets: CFP-FP \cite{sengupta2016frontal}, CPLFW \cite{cplfw}, AgeDB \cite{agedb}, and CALFW \cite{calfw}. }
\begin{tabular}{@{}c|ccc|cccc|c}
\toprule
 Method& MHMS& GFE& LGF& CFP-FP& CPLFW& AgeDB& CALFW& AVG\\
    \midrule
     $\uppercase\expandafter{\romannumeral1}$&$\checkmark$& &&$\underline{98.64}$&$\underline{93.07}$& 98.12& $\underline{96.18}$&96.50\\
 $\uppercase\expandafter{\romannumeral2}$&& $\checkmark$&& 98.27&92.08& $\textbf{98.28}$ & 95.45&96.02\\
 $\uppercase\expandafter{\romannumeral3}$&$\checkmark$& $\checkmark$&& 98.59 &$\textbf{93.42}$&$\underline{98.22}$& 96.10&$\underline{96.58}$ \\
    $\textbf{LGAF}$  &$\checkmark$& $\checkmark$&$\checkmark$&$\textbf{98.77}$&$\textbf{93.42}$& $\textbf{98.28}$ & $\textbf{96.22}$&$\textbf{96.67}$\\
    \bottomrule
  \end{tabular}

\label{tab:HQ ablation}
\end{subtable}

\begin{subtable}[t]{1\linewidth}
\centering
\subcaption{Closed set retrieval (Rank-1, Rank-3, Rank-5) on the TinyFace is reported.}
     \begin{tabular}{@{}c|ccc|ccc}
\toprule
 \multirow{2}{*}{Method}  & \multirow{2}{*}{MHMS}& \multirow{2}{*}{GFE}& \multirow{2}{*}{LGF}& \multicolumn{3}{c}{TinyFace}\\
 & & & & Rank-$1$& Rank-$3$&Rank-$5$\\
    \midrule
     $\uppercase\expandafter{\romannumeral1}$&$\checkmark$& && 67.84& 70.39&71.24\\
 $\uppercase\expandafter{\romannumeral2}$&& $\checkmark$&& 62.07*& 65.72*&67.19*\\
 $\uppercase\expandafter{\romannumeral3}$&$\checkmark$& $\checkmark$&& $\underline{68.03}$& $\underline{70.55}$&$\underline{71.35}$\\
    $\textbf{LGAF}$  &$\checkmark$& $\checkmark$&$\checkmark$& $\textbf{68.35}$& $\textbf{70.68}$&$\textbf{71.59}$\\
    \bottomrule
  \end{tabular}

 \label{tab:LQ ablation}
\end{subtable}

\end{table}

\subsubsection{Effects of the MHMS, the GFE, and the LGF} 
To investigate the effectiveness of our proposed module, CFP-FP, CPLFW, AgeDB, CALFW, and TinyFace datasets are used for performance evaluation. In Method $\uppercase\expandafter{\romannumeral1}$ and $\uppercase\expandafter{\romannumeral2}$ of Tab.\ref{tab:HQ ablation}, we explore the effectiveness of the MHMS module and the GFE module. In particular, when only global features are extracted, the model degenerates to ArcFace. When extracting only local features (Method $\uppercase\expandafter{\romannumeral1}$),  the recognition rate of CFP-FP, CPLFW, and CALFW is better than only extracting global features (Method $\uppercase\expandafter{\romannumeral2}$). The performance of the MHMS and the GFE on CFP-FP and AgeDB datasets reveals that the MHMS network excels on the validation set with significant pose variations (CFP-FP), while the GFE network outperforms on the validation set with substantial age variations (AgeDB). This further proves that global features are more discriminative when facial regions are significantly deformed, while local features are more conducive to FR where part of the facial region is missing. For TinyFace (As shown in Tab.\ref{tab:LQ ablation}), closed-set ranking retrieval demonstrates a marked enhancement in the performance of Method $\uppercase\expandafter{\romannumeral1}$ compared to $\uppercase\expandafter{\romannumeral2}$. 

By disabling the LGF module, Method $\uppercase\expandafter{\romannumeral3}$ of Tab.\ref{tab:HQ ablation} combines both local and global features directly by adding them together and normalizing them to obtain new fused features.  However, compared with Method $\uppercase\expandafter{\romannumeral1}$ and $\uppercase\expandafter{\romannumeral2}$, Method $\uppercase\expandafter{\romannumeral3}$ improves the recognition rate by 0.19\%, 0.16\%, and 0.11\% respectively in closed-set ranking retrieval on TinyFace (as described in Tab.\ref{tab:LQ ablation}), but does not produce the best performance on CFP-FP, AgeDB or CALFW. These suggest that simply combining local and global features does not directly enhance model recognition capability; it may be attributed to introducing redundant information.

To effectively complement local and global information and improve the quality of face features, we propose an LGF module. To verify the effectiveness of the LGF module, the performance comparison of the LGAF after ablating the LGF module is shown in Tab.\ref{tab:HQ ablation} for Method $\uppercase\expandafter{\romannumeral3}$ and LGAF.
Experimental results verify the effectiveness of the LGF module, and the highest performance is hit in all four 1:1 verification datasets. The recognition rate of the LGAF on the validation set with pose changes (CFP-FP) is improved from 98.64\% to 98.77\%. It maintains comparable recognition rates as Method $\uppercase\expandafter{\romannumeral2}$ on validation sets with age variation (AgeDB) while achieving the best performance in all closed-set ranking retrieval for TinyFace.

\begin{table}[!ht]
\centering
\caption{Ablation of the MHMS module parameter $b$.}
 \begin{tabular}{@{}c|cccc|c@{}}
    \toprule
    &CFP-FP&CPLFW & AgeDB&CALFW&AVG\\
    \midrule
 b=2& 96.71& 93.33& $\textbf{98.38}$& 96.05&96.12\\
 $\textbf{b=4}$& $\textbf{98.77}$& $\underline{93.42}$& 98.28& $\textbf{96.22}$&$\textbf{96.67}$\\
 b=8& $\underline{98.69}$& 93.20& 98.12& 96.12&$\underline{96.53}$\\
 b=16& 96.77& $\textbf{93.62}$& $\underline{98.32}$& $\underline{96.13}$&96.21\\
    \bottomrule
  \end{tabular}
  
  \label{tab:b Ablation}
\end{table}

\subsubsection{The settings of parameter $b$} 
For parameter $b$ ablation, we analyze the performance of the LGAF with various $b$ values. The results are detailed in Tab.\ref{tab:b Ablation}. We set the parameter $b$ to 2, 4, 8, and 16. As the value of $b$ increases, the average performance of four verification sets initially improves and then decreases. When $b$ is set to 2, average performance hits its lowest point; when $b$ is 4, the average recognition rate achieves its peak at 96.67\%. Consequently, we select $b$ to 4.

\begin{table}[!ht]
\centering
\caption{Performance comparison between different feature quality measures in the LGF module.}
\begin{tabular}{@{}c|cccc|c@{}}
    \toprule
    Method &CFP-FP&CPLFW & AgeDB&CALFW&AVG\\
     \midrule
 Baseline& $\underline{98.59}$& $\textbf{93.42}$& $\underline{98.22}$& $\underline{96.10}$&$\underline{96.58}$\\
     Entropy&$\underline{98.59}$& $\underline{93.05}$& 98.08&96.08&96.45\\
 $\textbf{Energy}$& $\textbf{98.77}$& $\textbf{93.42}$& $\textbf{98.28}$ & $\textbf{96.22}$&$\textbf{96.67}$\\
     \bottomrule
  \end{tabular}
  
  \label{tab:Ablation of feature quality measure}
  
\end{table}
\begin{table}[!ht]
\centering
\caption{Performance comparison between different feature fusion methods in the LGF module.}
\begin{tabular}{@{}c|cccc|c@{}}
    \toprule
    Method &CFP-FP&CPLFW & AgeDB&CALFW&AVG\\
     \midrule
 CAT& 97.90& 92.25& $\underline{98.10}$& 95.95&96.05\\
     iAFF&$\underline{98.60}$& $\underline{92.98}$& 98.02&$\underline{96.18}$&$\underline{96.45}$\\
 $\textbf{ADD}$& $\textbf{98.77}$& $\textbf{93.42}$& $\textbf{98.28}$ & $\textbf{96.22}$&$\textbf{96.67}$\\
     \bottomrule
  \end{tabular}
  
  \label{tab:Ablation of fusion methods}
  
\end{table}

\subsubsection{Effects of different fusion modes in the LGF} 
Regarding the LGF module, we explore three feature quality measures and three ways of fusing local and global features, aiming to find a more appropriate fusion strategy.

We compare baseline, energy-based (i.e., feature norm), and entropy-based \cite{dan2023transface} feature quality measures. As the Tab.\ref{tab:Ablation of feature quality measure} shows the feature quality evaluation metric based on energy (feature norm) achieves the best performance on all four validation sets. This also demonstrates the effectiveness of the feature norm.

We evaluate three fusion approaches: (a) Concatenating local and global features with different attention;  (b) Using the adapted iAFF \cite{dai2021attentional} (a multi-feature fusion module) to achieve fusion between local and global features; (c) Adding local and global features with different attention. The average recognition rate on four 1:1 verification sets is used as a performance metric.
As shown in Tab.\ref{tab:Ablation of fusion methods}, the experimental results indicate that employing a complex fusion network does not help the model obtain the best recognition performance and adds excessive cost. While the adapted iAFF feature fusion module improves the average performance from 96.05\% to 96.45\%, it does not help the model achieve the highest average performance. The attention-based addition of features elevates the average performance from 96.45\% to 96.67\% and achieves the best performance on all four validation sets. Consequently, we choose the attention-based addition mode in the feature fusion module of the LGAF. This method not only helps the model achieve excellent recognition performance but also does not incur excessive costs.

\begin{table}[!ht]
\centering
\caption{The performance of the recent methods is compared on four 1:1 verification sets, and the average performance is reported in column 6. ArcFace loss and CosFace loss are selected as loss functions to evaluate the model performance. *=our evaluation of the released model.}
\begin{tabular}{l|c|c|cccc|c}
\toprule
   Method&Year &BackBone& CFP-FP& CPLFW& AgeDB& CALFW&  AVG\\
        \midrule
 CosFace (m = 0.4) \cite{wang2018cosface}&2018 &ResNet100& 98.13*& 92.67*& 98.00*& 95.95*&96.19*\\
    ArcFace (m = 0.5) \cite{deng2019arcface} &2019 &ResNet100&98.27& 92.08& $\textbf{98.28}$ &95.45&96.02\\
 MV-Softmax \cite{wang2020mis}& 2020 &ResNet100& 98.28& 92.83& 97.95& 96.10& 96.29\\
 URL \cite{shi2020towards}   &2020 &ResNet100& 98.64& -& -& -&-  \\
 HPDA \cite{huang2020curricularface}  &2020 &LS-CNN& -& 92.35& -& 95.90&-  \\
 MagFace \cite{meng2021magface}  &2021 &ResNet100& 98.46& 92.87& 98.17& 96.15&96.41\\
 AdaFace \cite{kim2022adaface}  &2022 &ResNet100& 98.49& $\textbf{93.53}$& 98.05& 96.08& 96.54\\
 CQA-Face \cite{wang2022cqa}  &2022 &ResNet100& 98.49& 93.0& -& 96.12&-  \\
 TransFace-B \cite{dan2023transface}  &2023 &ViTs& 98.39*& 92.92*& 98.00*& 96.20*&96.38\\
 UniFace \cite{zhou2023uniface}& 2023 &ResNet100& 98.63*& 93.28*& $\underline{98.20}$*& 96.12*&96.56\\
 \midrule
 LGAF (CosFace)& Ours &ResNet100& $\underline{98.76}$& $\underline{93.42}$& 98.12& $\textbf{96.27}$&$\underline{96.64}$\\
 LGAF (ArcFace)&Ours &ResNet100& $\textbf{98.77}$& $\underline{93.42}$& $\textbf{98.28}$ & $\underline{96.22}$&$\textbf{96.67}$\\
      \bottomrule
    \end{tabular}

    \label{tab: HQ comparison with SoTA methods on MS1MV2}
\end{table}

\begin{table}[!ht]
\centering
\caption{The performance of the recent methods is compared on TinyFace and SCFace. For TinyFace, closed-set ranking retrieval (Rank-1, Rank-3, and Rank-5) is reported. For SCFace,  closed-set ranking retrieval (Rank-1) at distances d1,d2, and d3 are evaluated. ArcFace loss and CosFace loss are selected as loss functions to evaluate the model performance. *=our evaluation of the released model.}
\begin{subtable}[h]{1\linewidth}
\centering
\subcaption{The training set is MS1MV2.}
  \begin{tabular}{@{}l|c|c|ccc|ccc}
    \toprule
    \multirow{2}{*}{Method}    & \multirow{2}{*}{Year}  &\multirow{2}{*}{BackBone}&  \multicolumn{3}{c}{SCFace} &\multicolumn{3}{c}{TinyFace} \\
    & &&   d1& d2&d3&Rank-$1$&Rank-$3$&Rank-$5$    \\
        \midrule
 CosFace (m = 0.4) \cite{wang2018cosface} &2018 &ResNet100&   41.23*& 92.77*&$\underline{99.85}$*&62.37*&  65.98*&67.60*  \\
    ArcFace (m = 0.5) \cite{deng2019arcface}  &2019 &ResNet100&   41.38*& 92.92*&99.53*&62.07*&  65.72*&67.19*  \\
 URL \cite{shi2020towards}    &2020 &ResNet100&   -& -&-&63.89&-&68.67    \\
 AdaFace \cite{kim2022adaface}   &2022 &ResNet100&   66.46*& $\textbf{98.92}$*&$\textbf{1.00}$*&$\underline{68.21}$&  $\underline{70.60}$*&71.57 \\
 TransFace-B \cite{dan2023transface}   &2023 &ViTs&   $\underline{66.77}$*& 97.69*&99.69*&63.09*&65.91*&67.38*  \\
 TransFace-L \cite{dan2023transface} &2023 &ViTs& 66.76*& 96.92*& $\underline{99.85}$*& 67.52*& 69.85*&71.00*\\
     \midrule
  LGAF (CosFace) &Ours &ResNet100&   $\textbf{70.15}$& $\textbf{98.92}$&$\textbf{1.00}$&68.16&  $\underline{70.60}$&$\textbf{71.67}$  \\
 LGAF (ArcFace) &Ours &ResNet100&   66.30& $\underline{98.76}$&$\textbf{1.00}$&$\textbf{68.35}$&  $\textbf{70.68}$&$\underline{71.59}$ \\
      \bottomrule
    \end{tabular}

    \label{tab: LQ comparison with SoTA methods on MS1MV2}

\end{subtable}

\begin{subtable}[h]{1\linewidth}
    \centering
    \subcaption{The training set is WebFace4M.}
    \begin{tabular}{l|c|c|ccc|ccc}
        \toprule
    \multirow{2}{*}{Method}    & \multirow{2}{*}{Year}  &\multirow{2}{*}{BackBone}&  \multicolumn{3}{c}{SCFace}&\multicolumn{3}{c}{TinyFace} \\
    & &&  d1& d2& d3&Rank-$1$&Rank-$3$&Rank-$5$   \\
        \midrule
 CosFace (m = 0.4) \cite{wang2018cosface} &2018 &ResNet100&  $\underline{83.92}$*& $\textbf{1.00}$*& $\textbf{1.00}$*&70.82*&  73.20*&74.11*\\
 ArcFace (m = 0.5) \cite{deng2019arcface}  &2019 &ResNet100& 68.31*& 98.00*& 99.54*& 71.11& -&74.38 \\
 TransFace-B \cite{dan2023transface}   &2023 &ViTs& 80.15*& 98.77*& 99.85*& 62.77*& 66.09*&67.38*\\
 UniFace \cite{zhou2023uniface} &2023 &ResNet100& 81.69*& 99.69*& $\textbf{1.00}$*& 67.41*& 70.95*&73.44*\\
  \midrule
 LGAF (CosFace) &Ours &ResNet100&  $\textbf{87.38}$& $\underline{99.85}$& $\textbf{1.00}$&$\textbf{71.75}$&  $\textbf{74.22}$&$\textbf{74.92}$\\
 LGAF (ArcFace) &Ours &ResNet100& 78.15& $\textbf{1.00}$& $\textbf{1.00}$& $\underline{71.46}$& $\underline{73.58}$&$\underline{74.49}$\\
     \bottomrule
    \end{tabular}
            
              \label{tab: LQ comparison with SoTA methods on WebFace4M}
        \end{subtable}
               
  \label{tab:LQ comparison with SoTA methods}
\end{table}

\subsection{Comparison with SoTA Methods}

To compare with recent state-of-the-art methods, we conduct experiments on four 1:1 validation sets and two test sets to demonstrate the effectiveness of LGAF on low-quality test sets caused by various factors. We use ResNet100 as the backbone, employing ArcFace loss and CosFace loss as the loss functions. The comparison method is described as follows:

% \subsubsection{baseline Methods}

\textbf{Local feature representation network.} CQA-Face \cite{wang2022cqa} proposes to fully extract more discriminative information from the face feature map, enabling the model to maintain stable recognition even when key areas are occluded.
In addition, the HPDA \cite{wang2020hierarchical} extracts local features at different scales, but does not use an effective fusion strategy when fusing local and global features.

\textbf{Global feature representation network.} To align with the mainstream in the field of face recognition, we selecte CosFace and ArcFace \cite{deng2019arcface} as baseline. Based on the feature norm information, MagFace \cite{meng2021magface} and AdaFace \cite{kim2022adaface} contribute to learning a well-structured within-class feature distributions and highlighting difficult samples, respectively. MV-softmax \cite{wang2020mis} proposes to adaptively emphasize misclassified samples. URL \cite{shi2020towards} proposes to decouple facial features in terms of blur, occlusion, and pose to help the model improve generalization. UniFace \cite{zhou2023uniface} further constrains the separation of positive and negative examples by a learnable threshold in the training phase. The above models use a CNN-based backbone network to extract global features. Furthermore, TransFace \cite{dan2023transface} based on vision transformer (ViTs) proposes patch-level data augmentation and difficult sample mining strategies to alleviate the overfitting problem of ViT in FR.

% \textbf{Our method.} We propose an attention-based local and global feature fusion network that adaptively allocates attention based on feature quality, aiming to enhance the performance of low-quality FR.

As shown in Tab.\ref{tab: HQ comparison with SoTA methods on MS1MV2}, our proposed model achieves the best average recognition performance on four 1:1 validation sets that are primarily affected by pose variations (Category A in Fig.\ref{fig:Factors influencing FR}) or age variations (Category B in Fig.\ref{fig:Factors influencing FR}). Specifically, the LGAF achieves the best performance on three validation sets, ranks in the top two on four validation sets, and surpasses the state-of-the-art (SOTA) in terms of average recognition performance. For LGAF (ArcFace), the average recognition rate on four validation sets increases from 96.02\% of ArcFace to 96.67\%, and surpasses UniFace from 96.56\% to 96.67\%, reaching SOTA. For the LGAF using ArcFace loss as the loss function, it achieves SOTA on the CFP-FP and AgeDB validation sets, which are affected by pose and age variations, respectively. This further validates the effectiveness of our proposed model on low-quality image sets primarily influenced by factors in Categories A and B in Fig.\ref{fig:Factors influencing FR}. 
Compared with CosFace, the recognition performance of LGAF using CosFace loss as loss function on the four validation sets is improved from 98.13\% to 98.76\%, 92.67\% to 93.42\%, 98.00\% to 98.12\%, and 95.95\% to 96.27\%, respectively. Compared with TransFace on CALFW, the recognition rate of LGAF (CosFace) is improved from 96.20\% to 96.27\%. Furthermore, its recognition performance on CFP-FP, CPLFW, and the average performance reached the second position, only behind the LGAF (ArcFace) model.

For low-resolution test sets (Category C in Fig.\ref{fig:Factors influencing FR}), we use MS1MV2 as the training set and conduct closed-set retrieval on the SCFace and TinyFace. For SCFace, we evaluate Rank-1 retrieval performance at different distances (d1, d2, d3) to provide a comprehensive performance assessment. As shown in Tab.\ref{tab: LQ comparison with SoTA methods on MS1MV2}, the LGAF (CosFace) model outperforms the SOTA on SCFace. Compared to CosFace, it improves recognition performance by 28.92\%, 6.15\%, and 0.15\% at distances d1, d2, and d3, respectively. Compared to AdaFace, TransFace-B, and TransFace-L, the LGAF (CosFace) model improves d1 recognition performance by 3.69\%, 3.38\%, and 3.39\%, respectively. Additionally, using ArcFace loss as the loss function, the LGAF model achieves recognition improvements of 9.84\%, 2\%, and 0.46\% at distances d1, d2, and d3, respectively, compared to ArcFace.

Moreover, in closed-set retrieval on TinyFace, we compare Rank-1, Rank-3, and Rank-5 recognition rates. The LGAF demonstrates superior recognition performance. Specifically, the LGAF (CosFace) model improves recognition performance by 5.79\%, 4.62\%, and 4.07\% compared to CosFace. The LGAF (ArcFace) model improves recognition performance by 6.28\%, 4.96\%, and 4.4\% compared to ArcFace. Compared to AdaFace, which focuses on low-resolution samples, our proposed LGAF model achieves SOTA performance.

Furthermore, we investigate the performance of employing WebFace4M as the training set in contrast to the SOTA methods (as depicted in the Tab.\ref{tab: LQ comparison with SoTA methods on WebFace4M}). Compared to CosFace, LGAF (CosFace) improves by 3.46\% on d1, decreases by 0.15\% on d2, and is flat on d3. However, it achieves the SOTA on TinyFace, which improves the performance to 71.75\%, 74.22\%, and 74.92\%. Moreover, for the LGAF model with ArcFace as the loss function, the recognition performance is 9.84\%, 2\%, and 0.46\% superior to ArcFace on SCFace. And for TinyFace, LGAF (ArcFace) boosts the recognition rate by 4.05\%, 2.63\%, and 1.05\% compared to UniFace.

Experimental results show that the proposed LGAF network structure is compatible with different loss functions on different data sets, which excels in SoTA on low-resolution datasets while delivering exceptional performance on high-resolution validations.

\subsection{Analysis and discussion}

\begin{figure}[!t]
\centering
\includegraphics[scale=0.46]{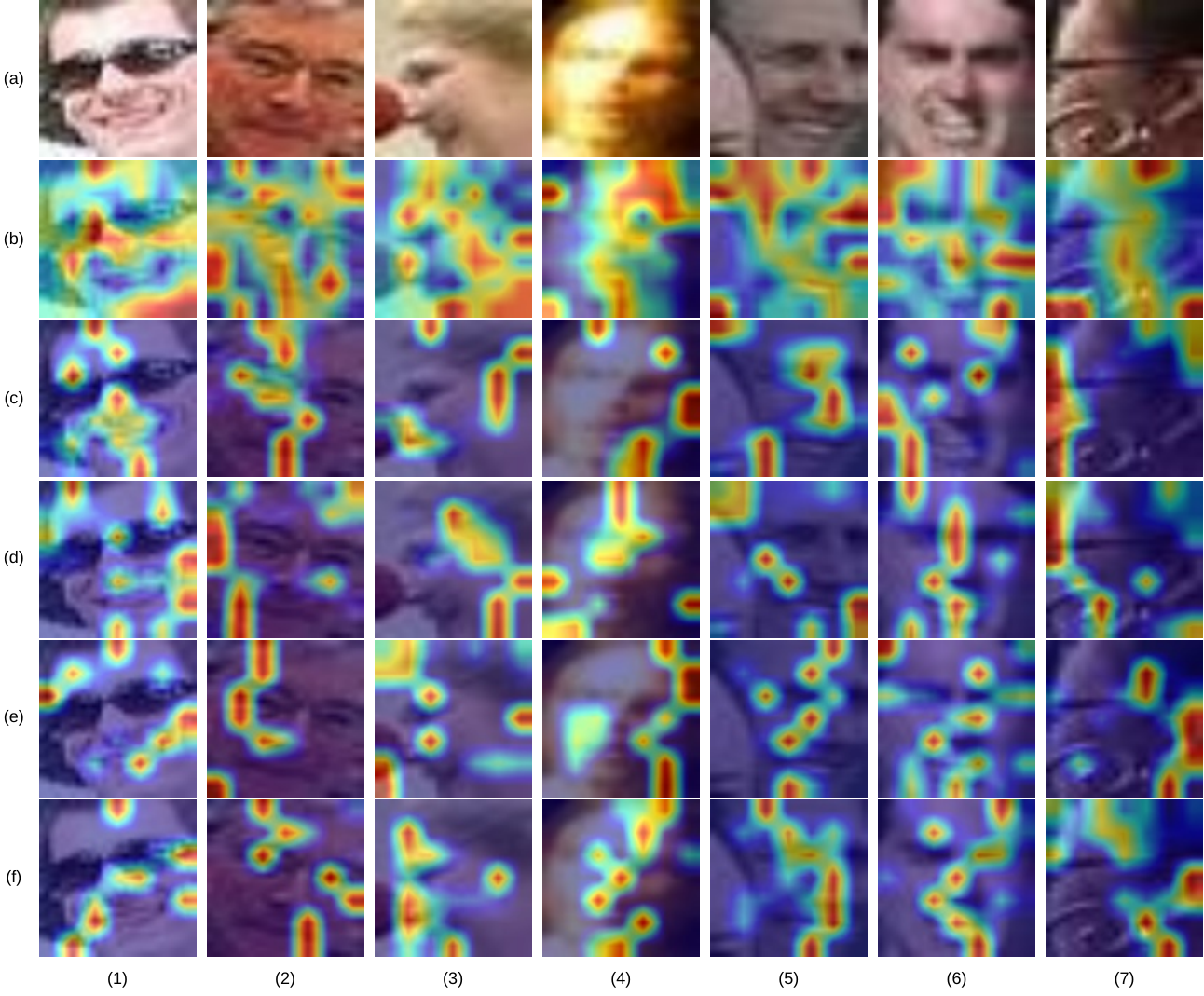}
\caption{Visualize local and global feature maps of seven low-quality face images in TinyFace. Row (a) represents the original image, row (b) represents the global feature map, and row (c) to (f) represents the local feature map extracted by the MHMS module.}
\label{fig:feature map on tinyface}
\end{figure}

For further analysis, low-quality face images on the TinyFace dataset were selected to visualize some local and global feature maps generated by the LGAF model. As shown in the Fig.\ref{fig:feature map on tinyface}, our selected images contain low-quality face images caused by partial face missing (occlusion, head pose variation), partial face deformation (expression variation, age variation), or joint influence of face missing and deformation (blur). We visualize the global feature map in row (b) and the feature maps of the four heads in the MHMS module in (c) to (f).
The global feature map mainly focuses on the structural regions composed of cheeks, nose, chin, etc., while the local feature map tends to focus on the corners of the mouth, eyebrows, nose tip and other regions, showing patchy distribution.

As proposed in Sec.\ref{LGF}, the contributions of local and global features to identity recognition are different for low-quality face images caused by different categories.
When the structural information is attenuated, such as the sunglasses occlusion of (1) in Fig.\ref{fig:feature map on tinyface}, the lack of the periocular region leads to the attenuation of the overall facial structural information, and focusing on the local region is more conducive to recognition. However, when the local deformation of the face occurs, such as the large deformation of the mouth shape in (9) of Fig.\ref{fig:feature map on tinyface}, the corresponding local feature map around the mouth will introduce noise. Therefore, local and global feature fusion through the LGF module can help the model achieve adaptive information complementarity.
The experimental results show that LGAF indeed shows good performance in low-quality face recognition tasks.

\section{Conclusion}

\label{sec:conclusion}
Based on previous research endeavors, we deeply analyze the causes of low-quality face images and summarize them into three categories: partial facial regions missing, substantial deformation in partial facial regions, and joint influence of the two above. We propose that in FR, low-quality samples affected by different categories have deviations in local and global feature quality, which affects the stability of FR. Therefore, we propose the LGAF network that fuses local and global features. The MHMS module and GFE module are employed to extract effective local and global features, and the local and global features are fused based on attention in the LGF module. Through effective information supplementation, more discriminative and high-quality facial features are generated. Experimental results show that LGAF achieves the average performance of SOTA on four 1:1 validation sets, and achieves the best performance on TinyFace, which is collected from real-world low-quality face images, and the SCFace dataset, which is collected using commercial cameras. This demonstrates the effectiveness and stability of our proposed model for low-quality face recognition.

\textbf{Limitations and future work.} Our proposed fusion of local and global features is a feature-level fusion strategy, where the model is based on the assumption that the local or global features that the sample has are correct and identifiable. However, there are unidentifiable and mislabeled samples in the training data. These samples introduce noise to the model, causing the model to learn incorrect information. Future work can consider adding a sample screening strategy in the training stage to reduce the model's attention to unidentifiable or mislabeled samples and enhance its focus on difficult samples.

 \bibliographystyle{style} 
 \bibliography{reference}

\begin{thebibliography}{41}
\expandafter\ifx\csname natexlab\endcsname\relax\def\natexlab#1{#1}\fi
\providecommand{\url}[1]{\texttt{#1}}
\providecommand{\href}[2]{#2}
\providecommand{\path}[1]{#1}
\providecommand{\DOIprefix}{doi:}
\providecommand{\ArXivprefix}{arXiv:}
\providecommand{\URLprefix}{URL: }
\providecommand{\Pubmedprefix}{pmid:}
\providecommand{\doi}[1]{\href{http://dx.doi.org/#1}{\path{#1}}}
\providecommand{\Pubmed}[1]{\href{pmid:#1}{\path{#1}}}
\providecommand{\bibinfo}[2]{#2}
\ifx\xfnm\relax \def\xfnm[#1]{\unskip,\space#1}\fi
%Type = Inproceedings
\bibitem[{Parde et~al.(2017)Parde, Castillo, Hill, Colon, Sankaranarayanan, Chen, and O’Toole}]{parde2016deep}
\bibinfo{author}{C.~J. Parde}, \bibinfo{author}{C.~Castillo}, \bibinfo{author}{M.~Q. Hill}, \bibinfo{author}{Y.~I. Colon}, \bibinfo{author}{S.~Sankaranarayanan}, \bibinfo{author}{J.-C. Chen}, \bibinfo{author}{A.~J. O’Toole},
\newblock \bibinfo{title}{Face and image representation in deep cnn features},
\newblock in: \bibinfo{booktitle}{2017 12th IEEE International Conference on Automatic Face \& Gesture Recognition (FG 2017)}, \bibinfo{year}{2017}, pp. \bibinfo{pages}{673--680}. \DOIprefix\doi{10.1109/FG.2017.85}.
%Type = Article
\bibitem[{Trigueros et~al.(2018)Trigueros, Meng, and Hartnett}]{trigueros2018face}
\bibinfo{author}{D.~S. Trigueros}, \bibinfo{author}{L.~Meng}, \bibinfo{author}{M.~Hartnett},
\newblock \bibinfo{title}{Face recognition: From traditional to deep learning methods},
\newblock \bibinfo{journal}{arXiv preprint arXiv:1811.00116}  (\bibinfo{year}{2018}).
%Type = Article
\bibitem[{Yang et~al.(2023)Yang, Wang, Huang, Xiao, Liang, Han, and Zou}]{YANG2023HeadPose}
\bibinfo{author}{J.~Yang}, \bibinfo{author}{Z.~Wang}, \bibinfo{author}{B.~Huang}, \bibinfo{author}{J.~Xiao}, \bibinfo{author}{C.~Liang}, \bibinfo{author}{Z.~Han}, \bibinfo{author}{H.~Zou},
\newblock \bibinfo{title}{Headpose-softmax: Head pose adaptive curriculum learning loss for deep face recognition},
\newblock \bibinfo{journal}{Pattern Recognition} \bibinfo{volume}{140} (\bibinfo{year}{2023}) \bibinfo{pages}{109552}. \URLprefix \url{https://www.sciencedirect.com/science/article/pii/S0031320323002522}. \DOIprefix\doi{https://doi.org/10.1016/j.patcog.2023.109552}.
%Type = Article
\bibitem[{Huang et~al.(2024)Huang, Rahardjo, Shiue, and Chen}]{HUANG2024110574}
\bibinfo{author}{Y.-C. Huang}, \bibinfo{author}{D.~A.~B. Rahardjo}, \bibinfo{author}{R.-H. Shiue}, \bibinfo{author}{H.~H. Chen},
\newblock \bibinfo{title}{Masked face recognition using domain adaptation},
\newblock \bibinfo{journal}{Pattern Recognition} \bibinfo{volume}{153} (\bibinfo{year}{2024}) \bibinfo{pages}{110574}. \URLprefix \url{https://www.sciencedirect.com/science/article/pii/S003132032400325X}. \DOIprefix\doi{https://doi.org/10.1016/j.patcog.2024.110574}.
%Type = Inproceedings
\bibitem[{Wang et~al.(2024)Wang, Sanchez, and Li}]{Wang2024Age}
\bibinfo{author}{H.~Wang}, \bibinfo{author}{V.~Sanchez}, \bibinfo{author}{C.-T. Li},
\newblock \bibinfo{title}{Cross-age contrastive learning for age-invariant face recognition},
\newblock in: \bibinfo{booktitle}{ICASSP 2024 - 2024 IEEE International Conference on Acoustics, Speech and Signal Processing (ICASSP)}, \bibinfo{year}{2024}, pp. \bibinfo{pages}{4600--4604}. \DOIprefix\doi{10.1109/ICASSP48485.2024.10445859}.
%Type = Article
\bibitem[{Zhang et~al.(2024)Zhang, Zheng, Li, Gao, and Lu}]{ZHANG2024Coupled}
\bibinfo{author}{K.~Zhang}, \bibinfo{author}{D.~Zheng}, \bibinfo{author}{J.~Li}, \bibinfo{author}{X.~Gao}, \bibinfo{author}{J.~Lu},
\newblock \bibinfo{title}{Coupled discriminative manifold alignment for low-resolution face recognition},
\newblock \bibinfo{journal}{Pattern Recognition} \bibinfo{volume}{147} (\bibinfo{year}{2024}) \bibinfo{pages}{110049}. \URLprefix \url{https://www.sciencedirect.com/science/article/pii/S003132032300746X}. \DOIprefix\doi{https://doi.org/10.1016/j.patcog.2023.110049}.
%Type = Article
\bibitem[{Zhao et~al.(2017)Zhao, Xiong, Karlekar~Jayashree, Li, Zhao, Wang, Sugiri~Pranata, Shengmei~Shen, Yan, and Feng}]{zhao2017dual}
\bibinfo{author}{J.~Zhao}, \bibinfo{author}{L.~Xiong}, \bibinfo{author}{P.~Karlekar~Jayashree}, \bibinfo{author}{J.~Li}, \bibinfo{author}{F.~Zhao}, \bibinfo{author}{Z.~Wang}, \bibinfo{author}{P.~Sugiri~Pranata}, \bibinfo{author}{P.~Shengmei~Shen}, \bibinfo{author}{S.~Yan}, \bibinfo{author}{J.~Feng},
\newblock \bibinfo{title}{Dual-agent gans for photorealistic and identity preserving profile face synthesis},
\newblock \bibinfo{journal}{Advances in neural information processing systems} \bibinfo{volume}{30} (\bibinfo{year}{2017}).
%Type = Inproceedings
\bibitem[{Shi and Jain(2021)}]{shi2021boosting}
\bibinfo{author}{Y.~Shi}, \bibinfo{author}{A.~K. Jain},
\newblock \bibinfo{title}{Boosting unconstrained face recognition with auxiliary unlabeled data},
\newblock in: \bibinfo{booktitle}{2021 IEEE/CVF Conference on Computer Vision and Pattern Recognition Workshops (CVPRW)}, \bibinfo{year}{2021}, pp. \bibinfo{pages}{2789--2798}. \DOIprefix\doi{10.1109/CVPRW53098.2021.00314}.
%Type = Article
\bibitem[{Song and Wang(2024)}]{CoReFace}
\bibinfo{author}{Y.~Song}, \bibinfo{author}{F.~Wang},
\newblock \bibinfo{title}{Coreface: Sample-guided contrastive regularization for deep face recognition},
\newblock \bibinfo{journal}{Pattern Recognition} \bibinfo{volume}{152} (\bibinfo{year}{2024}) \bibinfo{pages}{110483}. \URLprefix \url{https://www.sciencedirect.com/science/article/pii/S0031320324002346}. \DOIprefix\doi{https://doi.org/10.1016/j.patcog.2024.110483}.
%Type = Article
\bibitem[{Huang et~al.(2023)Huang, Wang, Wang, Jiang, Han, Lu, and Liang}]{HUANG2023PLFace}
\bibinfo{author}{B.~Huang}, \bibinfo{author}{Z.~Wang}, \bibinfo{author}{G.~Wang}, \bibinfo{author}{K.~Jiang}, \bibinfo{author}{Z.~Han}, \bibinfo{author}{T.~Lu}, \bibinfo{author}{C.~Liang},
\newblock \bibinfo{title}{Plface: Progressive learning for face recognition with mask bias},
\newblock \bibinfo{journal}{Pattern Recognition} \bibinfo{volume}{135} (\bibinfo{year}{2023}) \bibinfo{pages}{109142}. \URLprefix \url{https://www.sciencedirect.com/science/article/pii/S0031320322006227}. \DOIprefix\doi{https://doi.org/10.1016/j.patcog.2022.109142}.
%Type = Inproceedings
\bibitem[{Wang and Guo(2022)}]{wang2022cqa}
\bibinfo{author}{Q.~Wang}, \bibinfo{author}{G.~Guo},
\newblock \bibinfo{title}{Cqa-face: Contrastive quality-aware attentions for face recognition},
\newblock in: \bibinfo{booktitle}{Proceedings of the AAAI Conference on Artificial Intelligence}, volume~\bibinfo{volume}{36}, \bibinfo{year}{2022}, pp. \bibinfo{pages}{2504--2512}.
%Type = Article
\bibitem[{Su et~al.(2023)Su, Wang, Li, Gao, and Qiao}]{Hybrid}
\bibinfo{author}{W.~Su}, \bibinfo{author}{Y.~Wang}, \bibinfo{author}{K.~Li}, \bibinfo{author}{P.~Gao}, \bibinfo{author}{Y.~Qiao},
\newblock \bibinfo{title}{Hybrid token transformer for deep face recognition},
\newblock \bibinfo{journal}{Pattern Recognition} \bibinfo{volume}{139} (\bibinfo{year}{2023}) \bibinfo{pages}{109443}. \URLprefix \url{https://www.sciencedirect.com/science/article/pii/S0031320323001437}. \DOIprefix\doi{https://doi.org/10.1016/j.patcog.2023.109443}.
%Type = Inproceedings
\bibitem[{He et~al.(2016)He, Zhang, Ren, and Sun}]{he2016deep}
\bibinfo{author}{K.~He}, \bibinfo{author}{X.~Zhang}, \bibinfo{author}{S.~Ren}, \bibinfo{author}{J.~Sun},
\newblock \bibinfo{title}{Deep residual learning for image recognition},
\newblock in: \bibinfo{booktitle}{2016 IEEE Conference on Computer Vision and Pattern Recognition (CVPR)}, \bibinfo{year}{2016}, pp. \bibinfo{pages}{770--778}. \DOIprefix\doi{10.1109/CVPR.2016.90}.
%Type = Inproceedings
\bibitem[{Howard et~al.(2019)Howard, Sandler, Chen, Wang, Chen, Tan, Chu, Vasudevan, Zhu, Pang, Adam, and Le}]{howard2019mobilenets}
\bibinfo{author}{A.~Howard}, \bibinfo{author}{M.~Sandler}, \bibinfo{author}{B.~Chen}, \bibinfo{author}{W.~Wang}, \bibinfo{author}{L.-C. Chen}, \bibinfo{author}{M.~Tan}, \bibinfo{author}{G.~Chu}, \bibinfo{author}{V.~Vasudevan}, \bibinfo{author}{Y.~Zhu}, \bibinfo{author}{R.~Pang}, \bibinfo{author}{H.~Adam}, \bibinfo{author}{Q.~Le},
\newblock \bibinfo{title}{Searching for mobilenetv3},
\newblock in: \bibinfo{booktitle}{2019 IEEE/CVF International Conference on Computer Vision (ICCV)}, \bibinfo{year}{2019}, pp. \bibinfo{pages}{1314--1324}. \DOIprefix\doi{10.1109/ICCV.2019.00140}.
%Type = Inproceedings
\bibitem[{Deng et~al.(2021)Deng, Guo, Yang, Lattas, and Zafeiriou}]{deng2021variational}
\bibinfo{author}{J.~Deng}, \bibinfo{author}{J.~Guo}, \bibinfo{author}{J.~Yang}, \bibinfo{author}{A.~Lattas}, \bibinfo{author}{S.~Zafeiriou},
\newblock \bibinfo{title}{Variational prototype learning for deep face recognition},
\newblock in: \bibinfo{booktitle}{Proceedings of the IEEE/CVF Conference on Computer Vision and Pattern Recognition}, \bibinfo{year}{2021}, pp. \bibinfo{pages}{11906--11915}.
%Type = Article
\bibitem[{Boutros et~al.(2022)Boutros, Damer, Kirchbuchner, and Kuijper}]{Fadi2022Self}
\bibinfo{author}{F.~Boutros}, \bibinfo{author}{N.~Damer}, \bibinfo{author}{F.~Kirchbuchner}, \bibinfo{author}{A.~Kuijper},
\newblock \bibinfo{title}{Self-restrained triplet loss for accurate masked face recognition},
\newblock \bibinfo{journal}{Pattern Recognition} \bibinfo{volume}{124} (\bibinfo{year}{2022}) \bibinfo{pages}{108473}. \URLprefix \url{https://www.sciencedirect.com/science/article/pii/S003132032100649X}. \DOIprefix\doi{https://doi.org/10.1016/j.patcog.2021.108473}.
%Type = Inproceedings
\bibitem[{Wang et~al.(2020)Wang, Wu, Zheng, and Guo}]{wang2020hierarchical}
\bibinfo{author}{Q.~Wang}, \bibinfo{author}{T.~Wu}, \bibinfo{author}{H.~Zheng}, \bibinfo{author}{G.~Guo},
\newblock \bibinfo{title}{Hierarchical pyramid diverse attention networks for face recognition},
\newblock in: \bibinfo{booktitle}{2020 IEEE/CVF Conference on Computer Vision and Pattern Recognition (CVPR)}, \bibinfo{year}{2020}, pp. \bibinfo{pages}{8323--8332}. \DOIprefix\doi{10.1109/CVPR42600.2020.00835}.
%Type = Article
\bibitem[{Wang and Guo(2019)}]{wang2019ls}
\bibinfo{author}{Q.~Wang}, \bibinfo{author}{G.~Guo},
\newblock \bibinfo{title}{Ls-cnn: Characterizing local patches at multiple scales for face recognition},
\newblock \bibinfo{journal}{IEEE Transactions on Information Forensics and Security} \bibinfo{volume}{15} (\bibinfo{year}{2019}) \bibinfo{pages}{1640--1653}.
%Type = Article
\bibitem[{Hu et~al.(2020)Hu, Shen, Albanie, Sun, and Wu}]{hu2018squeeze}
\bibinfo{author}{J.~Hu}, \bibinfo{author}{L.~Shen}, \bibinfo{author}{S.~Albanie}, \bibinfo{author}{G.~Sun}, \bibinfo{author}{E.~Wu},
\newblock \bibinfo{title}{Squeeze-and-excitation networks},
\newblock \bibinfo{journal}{IEEE Transactions on Pattern Analysis and Machine Intelligence} \bibinfo{volume}{42} (\bibinfo{year}{2020}) \bibinfo{pages}{2011--2023}. \DOIprefix\doi{10.1109/TPAMI.2019.2913372}.
%Type = Article
\bibitem[{Ranjan et~al.(2017)Ranjan, Castillo, and Chellappa}]{ranjan2017l2}
\bibinfo{author}{R.~Ranjan}, \bibinfo{author}{C.~D. Castillo}, \bibinfo{author}{R.~Chellappa},
\newblock \bibinfo{title}{L2-constrained softmax loss for discriminative face verification},
\newblock \bibinfo{journal}{arXiv preprint arXiv:1703.09507}  (\bibinfo{year}{2017}).
%Type = Inproceedings
\bibitem[{Meng et~al.(2021)Meng, Zhao, Huang, and Zhou}]{meng2021magface}
\bibinfo{author}{Q.~Meng}, \bibinfo{author}{S.~Zhao}, \bibinfo{author}{Z.~Huang}, \bibinfo{author}{F.~Zhou},
\newblock \bibinfo{title}{Magface: A universal representation for face recognition and quality assessment},
\newblock in: \bibinfo{booktitle}{2021 IEEE/CVF Conference on Computer Vision and Pattern Recognition (CVPR)}, \bibinfo{year}{2021}, pp. \bibinfo{pages}{14220--14229}. \DOIprefix\doi{10.1109/CVPR46437.2021.01400}.
%Type = Inproceedings
\bibitem[{Kim et~al.(2022)Kim, Jain, and Liu}]{kim2022adaface}
\bibinfo{author}{M.~Kim}, \bibinfo{author}{A.~K. Jain}, \bibinfo{author}{X.~Liu},
\newblock \bibinfo{title}{Adaface: Quality adaptive margin for face recognition},
\newblock in: \bibinfo{booktitle}{2022 IEEE/CVF Conference on Computer Vision and Pattern Recognition (CVPR)}, \bibinfo{year}{2022}, pp. \bibinfo{pages}{18729--18738}. \DOIprefix\doi{10.1109/CVPR52688.2022.01819}.
%Type = Article
\bibitem[{Gross et~al.(2010)Gross, Matthews, Cohn, Kanade, and Baker}]{gross2010multi}
\bibinfo{author}{R.~Gross}, \bibinfo{author}{I.~Matthews}, \bibinfo{author}{J.~Cohn}, \bibinfo{author}{T.~Kanade}, \bibinfo{author}{S.~Baker},
\newblock \bibinfo{title}{Multi-pie},
\newblock \bibinfo{journal}{Image and vision computing} \bibinfo{volume}{28} (\bibinfo{year}{2010}) \bibinfo{pages}{807--813}.
%Type = Article
\bibitem[{Langner et~al.(2010)Langner, Dotsch, Bijlstra, Wigboldus, Hawk, and Van~Knippenberg}]{langner2010presentation}
\bibinfo{author}{O.~Langner}, \bibinfo{author}{R.~Dotsch}, \bibinfo{author}{G.~Bijlstra}, \bibinfo{author}{D.~H. Wigboldus}, \bibinfo{author}{S.~T. Hawk}, \bibinfo{author}{A.~Van~Knippenberg},
\newblock \bibinfo{title}{Presentation and validation of the radboud faces database},
\newblock \bibinfo{journal}{Cognition and emotion} \bibinfo{volume}{24} (\bibinfo{year}{2010}) \bibinfo{pages}{1377--1388}.
%Type = Inproceedings
\bibitem[{Szegedy et~al.(2015)Szegedy, Liu, Jia, Sermanet, Reed, Anguelov, Erhan, Vanhoucke, and Rabinovich}]{szegedy2015going}
\bibinfo{author}{C.~Szegedy}, \bibinfo{author}{W.~Liu}, \bibinfo{author}{Y.~Jia}, \bibinfo{author}{P.~Sermanet}, \bibinfo{author}{S.~Reed}, \bibinfo{author}{D.~Anguelov}, \bibinfo{author}{D.~Erhan}, \bibinfo{author}{V.~Vanhoucke}, \bibinfo{author}{A.~Rabinovich},
\newblock \bibinfo{title}{Going deeper with convolutions},
\newblock in: \bibinfo{booktitle}{2015 IEEE Conference on Computer Vision and Pattern Recognition (CVPR)}, \bibinfo{year}{2015}, pp. \bibinfo{pages}{1--9}. \DOIprefix\doi{10.1109/CVPR.2015.7298594}.
%Type = Inproceedings
\bibitem[{Deng et~al.(2019)Deng, Guo, Xue, and Zafeiriou}]{deng2019arcface}
\bibinfo{author}{J.~Deng}, \bibinfo{author}{J.~Guo}, \bibinfo{author}{N.~Xue}, \bibinfo{author}{S.~Zafeiriou},
\newblock \bibinfo{title}{Arcface: Additive angular margin loss for deep face recognition},
\newblock in: \bibinfo{booktitle}{2019 IEEE/CVF Conference on Computer Vision and Pattern Recognition (CVPR)}, \bibinfo{year}{2019}, pp. \bibinfo{pages}{4685--4694}. \DOIprefix\doi{10.1109/CVPR.2019.00482}.
%Type = Article
\bibitem[{Zhu et~al.(2023)Zhu, Huang, Deng, Ye, Huang, Chen, Zhu, Yang, Du, Lu, and Zhou}]{zhu2023webface260m}
\bibinfo{author}{Z.~Zhu}, \bibinfo{author}{G.~Huang}, \bibinfo{author}{J.~Deng}, \bibinfo{author}{Y.~Ye}, \bibinfo{author}{J.~Huang}, \bibinfo{author}{X.~Chen}, \bibinfo{author}{J.~Zhu}, \bibinfo{author}{T.~Yang}, \bibinfo{author}{D.~Du}, \bibinfo{author}{J.~Lu}, \bibinfo{author}{J.~Zhou},
\newblock \bibinfo{title}{Webface260m: A benchmark for million-scale deep face recognition},
\newblock \bibinfo{journal}{IEEE Transactions on Pattern Analysis and Machine Intelligence} \bibinfo{volume}{45} (\bibinfo{year}{2023}) \bibinfo{pages}{2627--2644}. \DOIprefix\doi{10.1109/TPAMI.2022.3169734}.
%Type = Inproceedings
\bibitem[{Guo et~al.(2016)Guo, Zhang, Hu, He, and Gao}]{guo2016ms}
\bibinfo{author}{Y.~Guo}, \bibinfo{author}{L.~Zhang}, \bibinfo{author}{Y.~Hu}, \bibinfo{author}{X.~He}, \bibinfo{author}{J.~Gao},
\newblock \bibinfo{title}{Ms-celeb-1m: A dataset and benchmark for large-scale face recognition},
\newblock in: \bibinfo{booktitle}{Computer Vision--ECCV 2016: 14th European Conference, Amsterdam, The Netherlands, October 11-14, 2016, Proceedings, Part III 14}, \bibinfo{organization}{Springer}, \bibinfo{year}{2016}, pp. \bibinfo{pages}{87--102}.
%Type = Inproceedings
\bibitem[{Sengupta et~al.(2016)Sengupta, Chen, Castillo, Patel, Chellappa, and Jacobs}]{sengupta2016frontal}
\bibinfo{author}{S.~Sengupta}, \bibinfo{author}{J.-C. Chen}, \bibinfo{author}{C.~Castillo}, \bibinfo{author}{V.~M. Patel}, \bibinfo{author}{R.~Chellappa}, \bibinfo{author}{D.~W. Jacobs},
\newblock \bibinfo{title}{Frontal to profile face verification in the wild},
\newblock in: \bibinfo{booktitle}{2016 IEEE winter conference on applications of computer vision (WACV)}, \bibinfo{organization}{IEEE}, \bibinfo{year}{2016}, pp. \bibinfo{pages}{1--9}.
%Type = Article
\bibitem[{Zheng and Deng(2018)}]{cplfw}
\bibinfo{author}{T.~Zheng}, \bibinfo{author}{W.~Deng},
\newblock \bibinfo{title}{Cross-{Pose} {LFW}: A database for studying cross-pose face recognition in unconstrained environments},
\newblock \bibinfo{journal}{Beijing University of Posts and Telecommunications, Tech. Rep} \bibinfo{volume}{5} (\bibinfo{year}{2018}) \bibinfo{pages}{7}.
%Type = Inproceedings
\bibitem[{Moschoglou et~al.(2017)Moschoglou, Papaioannou, Sagonas, Deng, Kotsia, and Zafeiriou}]{agedb}
\bibinfo{author}{S.~Moschoglou}, \bibinfo{author}{A.~Papaioannou}, \bibinfo{author}{C.~Sagonas}, \bibinfo{author}{J.~Deng}, \bibinfo{author}{I.~Kotsia}, \bibinfo{author}{S.~Zafeiriou},
\newblock \bibinfo{title}{Agedb: The first manually collected, in-the-wild age database},
\newblock in: \bibinfo{booktitle}{2017 IEEE Conference on Computer Vision and Pattern Recognition Workshops (CVPRW)}, \bibinfo{year}{2017}, pp. \bibinfo{pages}{1997--2005}. \DOIprefix\doi{10.1109/CVPRW.2017.250}.
%Type = Article
\bibitem[{Zheng et~al.(2017)Zheng, Deng, and Hu}]{calfw}
\bibinfo{author}{T.~Zheng}, \bibinfo{author}{W.~Deng}, \bibinfo{author}{J.~Hu},
\newblock \bibinfo{title}{Cross-age {LFW:} {A} database for studying cross-age face recognition in unconstrained environments},
\newblock \bibinfo{journal}{CoRR} \bibinfo{volume}{abs/1708.08197} (\bibinfo{year}{2017}). \URLprefix \url{http://arxiv.org/abs/1708.08197}. \href{http://arxiv.org/abs/1708.08197}{{\tt arXiv:1708.08197}}.
%Type = Inproceedings
\bibitem[{Cheng et~al.(2019)Cheng, Zhu, and Gong}]{cheng2019low}
\bibinfo{author}{Z.~Cheng}, \bibinfo{author}{X.~Zhu}, \bibinfo{author}{S.~Gong},
\newblock \bibinfo{title}{Low-resolution face recognition},
\newblock in: \bibinfo{booktitle}{Computer Vision--ACCV 2018: 14th Asian Conference on Computer Vision, Perth, Australia, December 2--6, 2018, Revised Selected Papers, Part III 14}, \bibinfo{organization}{Springer}, \bibinfo{year}{2019}, pp. \bibinfo{pages}{605--621}.
%Type = Article
\bibitem[{Grgic et~al.(2011)Grgic, Delac, and Grgic}]{2011SCface}
\bibinfo{author}{M.~Grgic}, \bibinfo{author}{K.~Delac}, \bibinfo{author}{S.~Grgic},
\newblock \bibinfo{title}{Scface – surveillance cameras face database},
\newblock \bibinfo{journal}{Multimedia Tools \& Applications} \bibinfo{volume}{51} (\bibinfo{year}{2011}) \bibinfo{pages}{863--879}.
%Type = Inproceedings
\bibitem[{Dan et~al.(2023)Dan, Liu, Xie, Deng, Xie, Xie, and Sun}]{dan2023transface}
\bibinfo{author}{J.~Dan}, \bibinfo{author}{Y.~Liu}, \bibinfo{author}{H.~Xie}, \bibinfo{author}{J.~Deng}, \bibinfo{author}{H.~Xie}, \bibinfo{author}{X.~Xie}, \bibinfo{author}{B.~Sun},
\newblock \bibinfo{title}{Transface: Calibrating transformer training for face recognition from a data-centric perspective},
\newblock in: \bibinfo{booktitle}{Proceedings of the IEEE/CVF International Conference on Computer Vision}, \bibinfo{year}{2023}, pp. \bibinfo{pages}{20642--20653}.
%Type = Inproceedings
\bibitem[{Dai et~al.(2021)Dai, Gieseke, Oehmcke, Wu, and Barnard}]{dai2021attentional}
\bibinfo{author}{Y.~Dai}, \bibinfo{author}{F.~Gieseke}, \bibinfo{author}{S.~Oehmcke}, \bibinfo{author}{Y.~Wu}, \bibinfo{author}{K.~Barnard},
\newblock \bibinfo{title}{Attentional feature fusion},
\newblock in: \bibinfo{booktitle}{Proceedings of the IEEE/CVF winter conference on applications of computer vision}, \bibinfo{year}{2021}, pp. \bibinfo{pages}{3560--3569}.
%Type = Inproceedings
\bibitem[{Wang et~al.(2018)Wang, Wang, Zhou, Ji, Gong, Zhou, Li, and Liu}]{wang2018cosface}
\bibinfo{author}{H.~Wang}, \bibinfo{author}{Y.~Wang}, \bibinfo{author}{Z.~Zhou}, \bibinfo{author}{X.~Ji}, \bibinfo{author}{D.~Gong}, \bibinfo{author}{J.~Zhou}, \bibinfo{author}{Z.~Li}, \bibinfo{author}{W.~Liu},
\newblock \bibinfo{title}{Cosface: Large margin cosine loss for deep face recognition},
\newblock in: \bibinfo{booktitle}{2018 IEEE/CVF Conference on Computer Vision and Pattern Recognition}, \bibinfo{year}{2018}, pp. \bibinfo{pages}{5265--5274}. \DOIprefix\doi{10.1109/CVPR.2018.00552}.
%Type = Inproceedings
\bibitem[{Wang et~al.(2020)Wang, Zhang, Wang, Fu, Shi, and Mei}]{wang2020mis}
\bibinfo{author}{X.~Wang}, \bibinfo{author}{S.~Zhang}, \bibinfo{author}{S.~Wang}, \bibinfo{author}{T.~Fu}, \bibinfo{author}{H.~Shi}, \bibinfo{author}{T.~Mei},
\newblock \bibinfo{title}{{Mis-classified} vector guided softmax loss for face recognition},
\newblock in: \bibinfo{booktitle}{Proceedings of the AAAI Conference on Artificial Intelligence}, volume~\bibinfo{volume}{34}, \bibinfo{year}{2020}, pp. \bibinfo{pages}{12241--12248}.
%Type = Inproceedings
\bibitem[{Shi et~al.(2020)Shi, Yu, Sohn, Chandraker, and Jain}]{shi2020towards}
\bibinfo{author}{Y.~Shi}, \bibinfo{author}{X.~Yu}, \bibinfo{author}{K.~Sohn}, \bibinfo{author}{M.~Chandraker}, \bibinfo{author}{A.~K. Jain},
\newblock \bibinfo{title}{Towards universal representation learning for deep face recognition},
\newblock in: \bibinfo{booktitle}{Proceedings of the IEEE/CVF Conference on Computer Vision and Pattern Recognition}, \bibinfo{year}{2020}, pp. \bibinfo{pages}{6817--6826}.
%Type = Inproceedings
\bibitem[{Huang et~al.(2020)Huang, Wang, Tai, Liu, Shen, Li, Li, and Huang}]{huang2020curricularface}
\bibinfo{author}{Y.~Huang}, \bibinfo{author}{Y.~Wang}, \bibinfo{author}{Y.~Tai}, \bibinfo{author}{X.~Liu}, \bibinfo{author}{P.~Shen}, \bibinfo{author}{S.~Li}, \bibinfo{author}{J.~Li}, \bibinfo{author}{F.~Huang},
\newblock \bibinfo{title}{Curricularface: Adaptive curriculum learning loss for deep face recognition},
\newblock in: \bibinfo{booktitle}{2020 IEEE/CVF Conference on Computer Vision and Pattern Recognition (CVPR)}, \bibinfo{year}{2020}, pp. \bibinfo{pages}{5900--5909}. \DOIprefix\doi{10.1109/CVPR42600.2020.00594}.
%Type = Inproceedings
\bibitem[{Zhou et~al.(2023)Zhou, Jia, Li, Shen, and Duan}]{zhou2023uniface}
\bibinfo{author}{J.~Zhou}, \bibinfo{author}{X.~Jia}, \bibinfo{author}{Q.~Li}, \bibinfo{author}{L.~Shen}, \bibinfo{author}{J.~Duan},
\newblock \bibinfo{title}{Uniface: Unified cross-entropy loss for deep face recognition},
\newblock in: \bibinfo{booktitle}{Proceedings of the IEEE/CVF International Conference on Computer Vision (ICCV)}, \bibinfo{year}{2023}, pp. \bibinfo{pages}{20730--20739}.

\end{thebibliography}




% %% For numbered reference style

%   \textit{\LaTeX: a document preparation system},
%   Addison Wesley, Massachusetts,

% \end{thebibliography}
\end{document}